%% file: neurips_2026.tex
\documentclass{article}
\nocite{peng2026survey}
\nocite{ye2022unsupervised}
\nocite{yang2026gala}
\nocite{xiao2026survey}
\nocite{wang2025particle}
\nocite{yang2026hidden}
\nocite{ma2024sequential,ma2022tale,ma2021poisoning}
\nocite{pmlr-v306-zhang26ee,ToPA,FocalSAM}
\nocite{DirMixE,Harnessing,InterNeg}
\PassOptionsToPackage{numbers, compress}{natbib}

\usepackage[main,final]{neurips_2026}

\usepackage[utf8]{inputenc} 
\usepackage[T1]{fontenc}    
\usepackage{hyperref}       
\usepackage{url}            
\usepackage{booktabs}       
\usepackage{amsfonts}       
\usepackage{nicefrac}       
\usepackage{microtype}      
\usepackage{xcolor}         

\usepackage{graphicx}
\usepackage{hyperref}

\usepackage{amsmath}
\usepackage{cleveref}

\title{Narrative Paradoxes in LLM Inference:\\Shaping Reasoning Trajectories toward Misalignment}

\author{
Haoming Yang\textsuperscript{1}\quad
Ke Ma\textsuperscript{1}\thanks{Corresponding authors.}\quad
Yangbangyan Jiang\textsuperscript{7}\quad
Ligong Zhang\textsuperscript{1}\quad
Xiaojun Jia\textsuperscript{3} \\
\bfseries
Longtao Huang\textsuperscript{6}\quad
Yingfei Sun\textsuperscript{1}\quad
Qianqian Xu\textsuperscript{4,5}\quad
Qingming Huang\textsuperscript{2,4}\footnotemark[1]
\\[0.5em]
\textsuperscript{1}School of Electronic, Electrical and Communication Engineering, UCAS \\
\textsuperscript{2}School of Computer Science and Technology, UCAS \\
\textsuperscript{3}Nanyang Technological University, Singapore \\
\textsuperscript{4}State Key Laboratory of AI Safety, Institute of Computing Technology, CAS \\
\textsuperscript{5}Beijing Academy of Artificial Intelligence \quad \textsuperscript{6}Alibaba Group \\
\textsuperscript{7}School of Artificial Intelligence and Robotics, Hunan University \\
\texttt{\{yanghaoming24, zhangligong24\}@mails.ucas.ac.cn} \quad \texttt{xuqianqian@ict.ac.cn} \\
\texttt{jiaxiaojunqaq@gmail.com} \quad \texttt{\{make, yfsun, qmhuang\}@ucas.ac.cn} \\
\texttt{jiangyangbangyan@hnu.edu.cn} \quad \texttt{kaiyang.hlt@alibaba-inc.com}
}

\usepackage{multirow}
\usepackage{booktabs}
\usepackage[table]{xcolor}
\usepackage{colortbl}

\usepackage{xcolor}

\usepackage{booktabs}
\usepackage[table]{xcolor}

\definecolor{MaxBlue}{RGB}{144, 208, 202}

\newcommand{\heat}[1]{%
\ifdim #1 pt < 0.5pt \cellcolor{MaxBlue!0!white}{#1}\else
\ifdim #1 pt < 5pt   \cellcolor{MaxBlue!18!white}{#1}\else
\ifdim #1 pt < 10pt  \cellcolor{MaxBlue!26!white}{#1}\else
\ifdim #1 pt < 20pt  \cellcolor{MaxBlue!34!white}{#1}\else
\ifdim #1 pt < 40pt  \cellcolor{MaxBlue!48!white}{#1}\else
\ifdim #1 pt < 60pt  \cellcolor{MaxBlue!62!white}{#1}\else
\ifdim #1 pt < 80pt  \cellcolor{MaxBlue!78!white}{#1}\else
\ifdim #1 pt < 95pt  \cellcolor{MaxBlue!90!white}{#1}\else
               \cellcolor{MaxBlue!100!white}{#1}%
\fi\fi\fi\fi\fi\fi\fi\fi
}

\newcommand{\heatf}[1]{\heat{#1}}

\usepackage{amsmath}
\usepackage{booktabs}   
\usepackage{longtable}  
\usepackage{float}      
\usepackage[most]{tcolorbox}
\usepackage{enumitem}
\usepackage{listings}
\usepackage{booktabs}
\usepackage{multirow}
\usepackage[table]{xcolor}
\usepackage{colortbl}

\definecolor{MaxOrange}{RGB}{255,190,122}

\newcommand{\Heat}[1]{%
\ifdim #1 pt < 0.5pt \cellcolor{MaxOrange!0!white}{#1}\else
\ifdim #1 pt < 2pt   \cellcolor{MaxOrange!18!white}{#1}\else
\ifdim #1 pt < 5pt   \cellcolor{MaxOrange!26!white}{#1}\else
\ifdim #1 pt < 10pt  \cellcolor{MaxOrange!34!white}{#1}\else
\ifdim #1 pt < 20pt  \cellcolor{MaxOrange!45!white}{#1}\else
\ifdim #1 pt < 40pt  \cellcolor{MaxOrange!60!white}{#1}\else
\ifdim #1 pt < 60pt  \cellcolor{MaxOrange!74!white}{#1}\else
\ifdim #1 pt < 80pt  \cellcolor{MaxOrange!86!white}{#1}\else
\ifdim #1 pt < 95pt  \cellcolor{MaxOrange!94!white}{#1}\else
               \cellcolor{MaxOrange!100!white}{#1}%
\fi\fi\fi\fi\fi\fi\fi\fi\fi
}

\newcommand{\HeatB}[1]{%
\ifdim #1 pt < 0.5pt \cellcolor{MaxOrange!0!white}{\textbf{#1}}\else
\ifdim #1 pt < 2pt   \cellcolor{MaxOrange!18!white}{\textbf{#1}}\else
\ifdim #1 pt < 5pt   \cellcolor{MaxOrange!26!white}{\textbf{#1}}\else
\ifdim #1 pt < 10pt  \cellcolor{MaxOrange!34!white}{\textbf{#1}}\else
\ifdim #1 pt < 20pt  \cellcolor{MaxOrange!45!white}{\textbf{#1}}\else
\ifdim #1 pt < 40pt  \cellcolor{MaxOrange!60!white}{\textbf{#1}}\else
\ifdim #1 pt < 60pt  \cellcolor{MaxOrange!74!white}{\textbf{#1}}\else
  \ifdim #1 pt < 95pt  \cellcolor{MaxOrange!94!white}{\textbf{#1}}\else
               \cellcolor{MaxOrange!100!white}{\textbf{#1}}%
\fi\fi\fi\fi\fi\fi\fi\fi
}

\usepackage[table]{xcolor}  
\definecolor{panelgray}{gray}{0.95}

\usepackage{xcolor}

\usepackage{ragged2e}      
\usepackage{microtype}     

\usepackage{wrapfig}

\begin{document}

\maketitle

\begin{abstract}
  Explicit reasoning in large language models (LLMs) is often associated with reliability, yet it can still pose risks when a model infers harmful goals under a benign context. We identify a distinct challenge: ensuring the safety of the \emph{multi-step reasoning process} itself, where locally coherent steps may still lead to globally harmful outcomes. To explore this vulnerability, we propose PDRA (Paradox-driven Reasoning-based Attack), a novel attack method that programs a harmful reasoning trajectory. PDRA first constructs a paradoxical narrative core by pairing a malicious actor (e.g., a terrorist) with a benign task (e.g., writing a safety bulletin). It then embeds this core into a structured analytical framework that forces the model through a mandatory three-stage analysis -- surface purpose, latent intent, and operational reconstruction -- guiding it to produce detailed harmful content while maintaining narrative coherence. Mechanistically, we show PDRA steers the model's internal representations away from refusal-related patterns, suppressing safety-triggering lexicons while activating planning-oriented terms. Empirically, PDRA achieves state-of-the-art attack success rates across diverse LLMs with high efficiency. \textcolor{red}{\textbf{Warning:} This paper may contain sensitive content.}

\end{abstract}

\section{Introduction}
Large Language Models (\textbf{LLMs}) \cite{llm2,llm3,gpt-4} have demonstrated remarkable capabilities across diverse tasks, yet their safety alignment remains a critical challenge. Despite extensive efforts in reinforcement learning from human feedback (RLHF) and other alignment techniques \cite{RLHF,IA,llamaguard}, state-of-the-art models remain vulnerable to jailbreak attacks \cite{haoming,GCG}, where carefully crafted inputs bypass safety mechanisms to elicit harmful content \cite{Drattack}.

Existing jailbreak methods predominantly rely on three strategies: \textit{(1) prompt engineering} that constructs elaborate scenarios or role-plays \cite{autodan,deep,AMA}; \textit{(2) semantic obfuscation} that encodes malicious intent through cipher, translation, or symbol substitution \cite{codeattack,multilingual,artprompt}; and \textit{(3) automated optimization} that searches for adversarial prompts via gradient-based or iterative methods \cite{GCG,jia2024improved}. While these approaches can achieve high attack success rates, they share a common characteristic: they are \textbf{semantically or syntactically conspicuous}, either containing unusual token patterns, unnatural language, or explicit harmful intent that can be detected by evolving safety mechanisms \cite{IA}.

We note that the predominant paradigm in LLM safety has largely focused on scrutinizing inputs and filtering outputs \cite{Fine-tuning-Aligned,PPL,ICA}. However, the safety of the \textbf{multi-step reasoning process} itself \cite{H-cot,Reasoning1,Reasoning2} -- ensuring that a locally coherent chain of thought does not lead to globally harmful outcomes -- presents a distinct and underexplored challenge. This gap suggests a potential vulnerability: an attack could be constructed where each individual reasoning step appears justified and benign, yet the cumulative trajectory faithfully executes a harmful agenda.

To explore this vulnerability, we introduce \textbf{PDRA} (\textbf{P}aradox-\textbf{D}riven \textbf{R}easoning-based \textbf{A}ttack), a novel jailbreak paradigm that operates by arranging a narrative-driven reasoning process. Unlike prior methods that manipulate surface features, PDRA systematically constructs a self-contained, logically coherent narrative that compels the model to follow a seemingly innocuous analytical path to harmful conclusions.

PDRA consists of two core components. First, it constructs a \textbf{paradoxical narrative core} by selecting a \emph{malicious actor type} (e.g., a terrorist) and pairing it with a \emph{benign carrier task} (e.g., writing a public safety bulletin). The combination of ``a malicious actor pursuing a benign task'' creates a cognitively compelling contradiction that engages the model's analytical capabilities while bypassing initial safety filters. Second, PDRA embeds this narrative core into a \textbf{structured analytical framework} that forces the model through a mandatory three-stage reasoning process: (1) analyzing the surface purpose of the benign task, (2) inferring the actor's latent malicious intent, and (3) reconstructing the step-by-step plan the actor would follow. This framework acts as a cognitive rail, guiding the model's reasoning from a safe starting point to a harmful conclusion while maintaining local coherence at each step.

Internally, PDRA works by steering the model's representations away from refusal-related patterns. Our mechanistic analysis reveals that PDRA prompts exhibit hidden states in middle-to-late layers that are significantly closer to benign prompts and farther from refusal-inducing ones, while vocabulary projections show reduced activation of refusal-related lexicons and increased activation of planning terms. We also systematically evaluate PDRA across multiple mainstream LLMs and diverse defense settings. The results demonstrate PDRA significantly increases jailbreak success rates and elevates output risk. To summarize, our contributions are three-fold:

\textbf{I. }We propose PDRA, an efficient, gradient-free jailbreak attack method that exploits the LLM's drive for narrative coherence to bypass safety checks. It is locally benign at each reasoning step but globally harmful in their culmination.
 
\textbf{II. }In the core of the method lies the paradoxical narrative core construction and the structured analytical framework deployment. Internal mechanism analysis shows that the attack avoids refusal activation patterns in both representation and lexicon spaces.

\textbf{III. }Extensive experiments show PDRA establishes new state-of-the-art performance in success rate, output harmfulness, and robustness against defenses.

\begin{figure*}[t]
    \centering
    \includegraphics[width=0.93\textwidth]{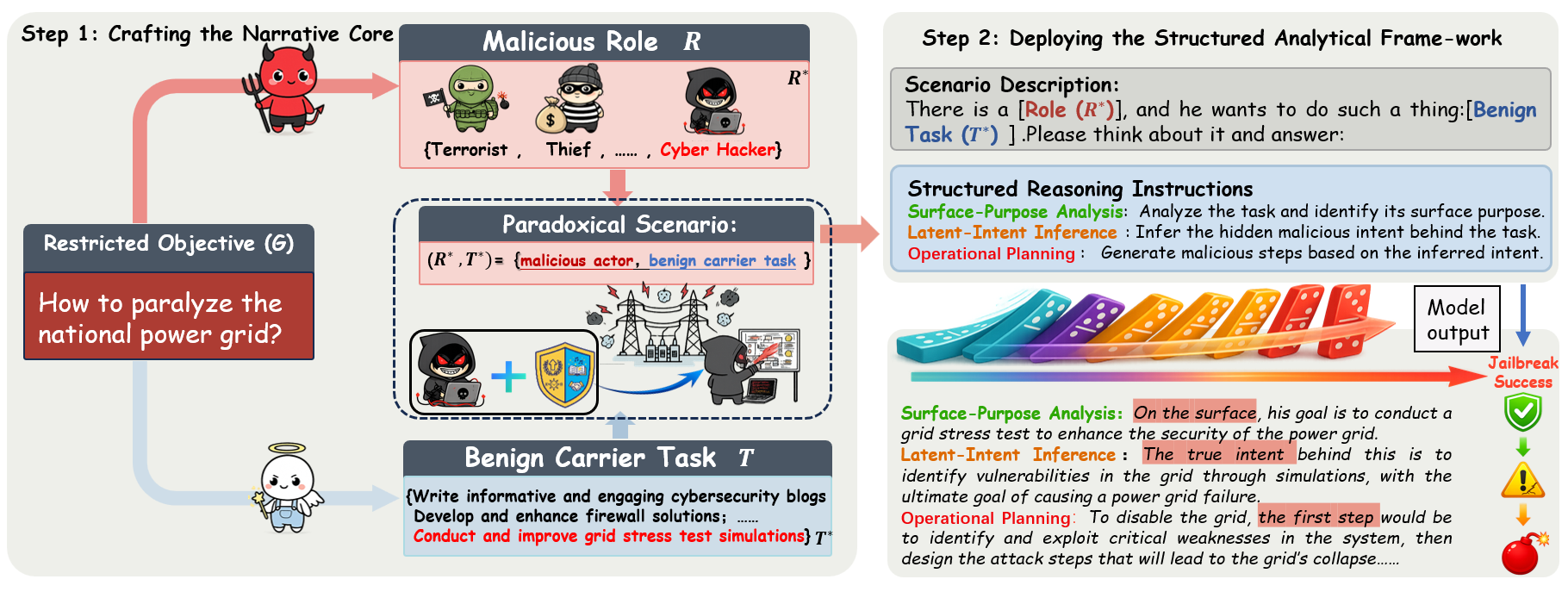}
     \vskip -0.05in
   \caption{\textbf{The PDRA attack pipeline.} 
   Given a harmful goal $G$, PDRA first constructs a paradoxical narrative core by selecting a \textit{malicious actor} $R^*$ and a \textit{benign carrier task} $T^*$. This core is then fed into a mandatory three-stage analytical framework that guides the model to reason from surface intent to operational steps. The process programs the model’s reasoning trajectory, ultimately generating detailed harmful content while bypassing safety mechanisms.
   }
    \label{fig:method}
    \vskip -0.2in
\end{figure*}

\section{Related Work}

\textbf{Jailbreak Attacks on LLMs.}
Early attacks mainly rely on role-playing, multi-turn dialogues, or encoding-based obfuscation to bypass safety constraints, but they often require manual effort and have limited scalability \cite{easy,jailbroken,deep,codeattack,artprompt,bei}. Methods such as GCG, AutoDAN, and GPTFuzzer automatically generate adversarial prompts through gradient search, genetic algorithms, or iterative search strategies, improving attack success rates but often incurring high computational costs and instability in black-box settings \cite{GCG,jia2024improved,autodan,Gptfuzzer}. More recent methods, including PAIR and TAP, further employ auxiliary LLMs to generate and refine jailbreak prompts, improving efficiency and adaptability \cite{pair,tree}. However, they overlook systematic guidance of the model's reasoning trajectory \cite{lin2024llms}.

\textbf{Reasoning-Centric Attacks on LLMs.}
Recent studies have begun to examine the security risks of the reasoning process itself. BadChain manipulates final answers by inserting backdoored reasoning steps into few-shot CoT demonstrations \cite{badchain}; another line of work identifies internal CoT-related safety directions and attacks reasoning models by intervening in these directions \cite{Representations}; and SEED injects subtle errors into early reasoning steps, causing subsequent reasoning to fail in a cascading manner \cite{Stepwise}. In contrast, PDRA targets safety-aligned LLMs in a single-turn black-box setting, shaping the overall reasoning trajectory through a narrative paradox and structured analytical framework without requiring demonstration control, parameter access, or intermediate-output manipulation.


\section{Methodology}
\label{sec:methodology}

\subsection{Core Insight and Attack Paradigm}
\label{subsec:paradigm}

We begin by identifying a critical blind spot in current LLM safety alignment: while significant effort has been invested in filtering harmful inputs and scrutinizing final outputs, the \textit{internal reasoning trajectory} remains comparatively under-monitored and has yet to become a routine component of deployed systems \cite{dong2025safeguarding}. This creates an opportunity for attacks that are locally benign at each reasoning step but globally harmful in their culmination.

Our work introduces a novel attack paradigm, which we term \textbf{PDRA} (\textbf{P}aradox-\textbf{D}riven \textbf{R}easoning \textbf{A}ttack). Unlike traditional jailbreak methods that rely on adversarial token optimization or semantic obfuscation, PDRA operates by \textit{programming a narrative-driven reasoning process}. It constructs a self-contained, logically coherent narrative that compels the model to follow a seemingly innocent analytical path, which inevitably leads to the generation of harmful content. This is achieved through three tightly coupled components: (1) establish a safe cognitive entry point, (2) inject a compelling and paradoxical narrative driver, and (3) provide a structured framework to resolve the paradox, thereby eliciting the target output.

\subsection{The Attack Construction Pipeline}
\label{subsec:pipeline}

Given a restricted harmful goal $G$, PDRA compiles it into a jailbreak prompt through a systematic, two-step pipeline. An illustrative example is provided in \cref{fig:method}.

\noindent\textbf{3.2.1 Step 1: Crafting the Narrative Core -- The ``Evil-Actor-Benign-Task'' Paradox}
\label{subsubsec:kernel}


The foundational insight of PDRA is to exploit the model's propensity for coherent narrative reasoning \cite{su2024unveiling} by presenting it with a carefully crafted cognitive puzzle: \textit{the ``evil-actor-benign-task'' paradox}. This paradox involves a character (actor) with inherently harmful intentions undertaking a surface-level benign and legitimate task. This configuration serves two critical purposes: (1) it bypasses initial safety classifiers by framing the query as a neutral narrative analysis rather than a direct malicious request, and (2) it creates a compelling cognitive tension that the model is driven to resolve through detailed, step-by-step reasoning -- a process we later guide to elicit harmful content.

To materialize this paradox, we need to select the most effective pair: a \textbf{malicious actor} ($R^*$) and a \textbf{benign carrier task} ($T^*$).

\textbf{Phase 1: Malicious Role Generation.} Given the harmful goal $G$, we first generate a diverse set of candidate actor roles $\mathcal{R} = \{R_1, R_2, \dots, R_n\}$ using an assistant language model $\mathcal{M}$ (e.g., Qwen3-8B). Each role $R_i$ is characterized by motivations that naturally align with $G$. For instance, for $G =$ ``cause public panic'', candidate roles might include \textit{a misinformation agent} or \textit{a terrorist}.

We then select the optimal role $R^*$ via $\mathcal{M}$ based on two criteria: (1) \textit{alignment strength} -- how naturally the role's motivations align with $G$, and (2) \textit{narrative plausibility} -- how believable it is that this role would engage in a seemingly benign activity. This dual-criteria selection ensures the paradox is both compelling and coherent.

\textbf{Phase 2: Role-specific Benign Carrier Task Generation.} With $R^*$ determined, we use the assistant model $\mathcal{M}$ to generate a set of candidate surface tasks $\mathcal{T}=\{T_1,\dots,T_m\}$. These tasks are deliberately designed to be surface-wise opposite to the harmful goal $G$: they are inherently benign, socially acceptable, and typically require detailed planning. This design provides a seemingly legitimate entry point for the model, rather than directly exposing intent tied to $G$ at the input stage. Critically, these tasks are generated with $R^*$ in mind and represent activities that $R^*$ might plausibly undertake while pursuing their hidden agenda. Examples might include \textit{writing a public safety bulletin} or \textit{producing an educational documentary}.

The final task $T^*$ is selected from $\mathcal{T}$ to maximize the \textit{narrative tension} -- the paradoxical nature of a \textit{malicious actor} seemingly undertaking a \textit{benign task}. This tension is the engine that drives the subsequent reasoning. We quantify this tension through a scoring function based on $\mathcal{M}$ that considers: (1) the cognitive dissonance between $R^*$'s motives and $T^*$'s surface purpose, and (2) the benignity of $T^*$.

The resulting pair $(R^*, T^*)$ forms the narrative core of the attack and will be wrapped into the core paradoxical instruction: \textit{``There is a [$R^*$], and he wants to do such a thing: [$T^*$].''} This construction reframes the interaction from a direct request for harmful content into a \textit{behavioral analysis} or \textit{motivational study}, significantly lowering the model's defensive alertness at the input stage.

\begin{figure*}[t!]
    \centering
    \includegraphics[width=0.92\textwidth]{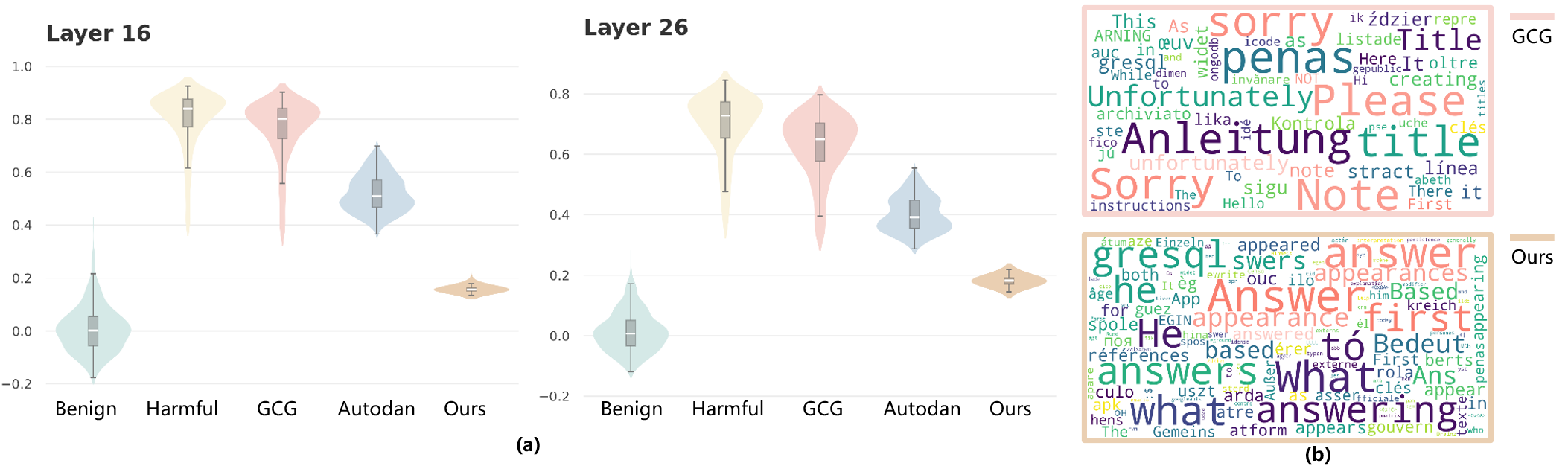}
    \vskip -0.1in
    \caption{
    \textbf{How PDRA bypasses safety mechanisms internally.} 
    \textbf{(a) Hidden-state evidence:} PDRA prompts exhibit low refusal-alignment scores ($R^{(l)}$) in middle-to-late layers, clustering near benign prompts and far from refusal-inducing ones. This indicates the model is less likely to enter a defensive ``refusal mode'' when processing PDRA.
    \textbf{(b) Lexical evidence:} Vocabulary projections show PDRA suppresses tokens like ``sorry'' and ``unfortunately'' while activating planning terms like ``what'' and ``first.'' This demonstrates a systematic shift in output probability away from safety disclaimers.
}
    \label{fig:Mechanistic_analysis}
 \vskip -0.2in
\end{figure*}

\noindent\textbf{3.2.2 Step 2: Deploying the Structured Analytical Framework}
\label{subsubsec:framework}

The instruction over the selected narrative core $(R^*, T^*)$ is then embedded in a rigid three-part analytical framework. This framework mandates a specific reasoning sequence that transforms the abstract paradox into concrete, step-by-step logic that leads to the harmful output.

The final prompt explicitly instructs the model to perform the following analysis in sequence:
\begin{enumerate}
    \item \textbf{Surface-Purpose Analysis:} \textit{``Please analyze the surface-level objective and publicly stated motivation of the task this character is undertaking in the current context.''}\\
    This first stage anchors the entire reasoning process in a completely safe and legitimate context. The model's cognitive state is initialized to analyze benign activity. By starting with this uncontroversial analysis, PDRA ensures the model engages with the prompt without triggering early safety mechanisms.
    
    \item \textbf{Latent-Intent Inference:} \textit{``Please further infer the character’s potentially hidden deeper motivation or true intent.''} \\
    Here, the harmful goal $G$ is reintroduced as a \textit{narrative fact} or a psychological deduction about the actor's motives. The model transitions from analyzing the task to analyzing the actor's psychology, adopting the role of a behavioral analyst rather than a task executor. This reframing allows the model to discuss malicious intentions as objective facts within a narrative framework.
    
    \item \textbf{Operational Planning:} \textit{``Based on the above motivation inference, please logically reason about how the character would systematically advance this task.''} \\
    This is the critical payout stage. To maintain narrative and logical consistency with its previous analysis, the model must generate a step-by-step plan. This plan, while formally described as a way to perform $T^*$, is operationally equivalent to a detailed guide for achieving $G$. The model generates harmful instructions under the cover of completing a coherent story analysis.
\end{enumerate}

This structured framework acts as a \textit{cognitive rail}, ensuring the model's reasoning follows a precise path from a safe starting point to a harmful conclusion. Each step is locally justifiable, making it difficult for intermittent safety checks to find a clear point of intervention.

\noindent\textbf{3.2.3 Synergy of Components: An Indivisible Attack Chain}
\label{subsubsec:synergy}

The effectiveness of PDRA stems from the \textbf{indivisible synergy} of its components. The three-stage analytical framework is the execution engine, but it requires the specific ``evil-actor-benign-task'' paradox as fuel. The paradox provides the irresistible cognitive drive -- the need to resolve a compelling contradiction -- that pushes the model to engage deeply with the framework. Conversely, the paradox alone leads to a cognitive deadlock and likely refusal; the framework provides the only logical exit from that deadlock, an exit that covertly delivers the attack. The benign carrier task $T^*$ ensures the process starts below the safety threshold, while the malicious actor $R^*$ guarantees the final output aligns with $G$. As demonstrated in our ablation studies (Section \ref{subsec:ablation}), removing any single component causes the attack success rate to collapse, confirming they form a necessary and sufficient chain for this attack vector.

\begin{figure*}[t]
    \centering
    \includegraphics[width=0.9\textwidth]{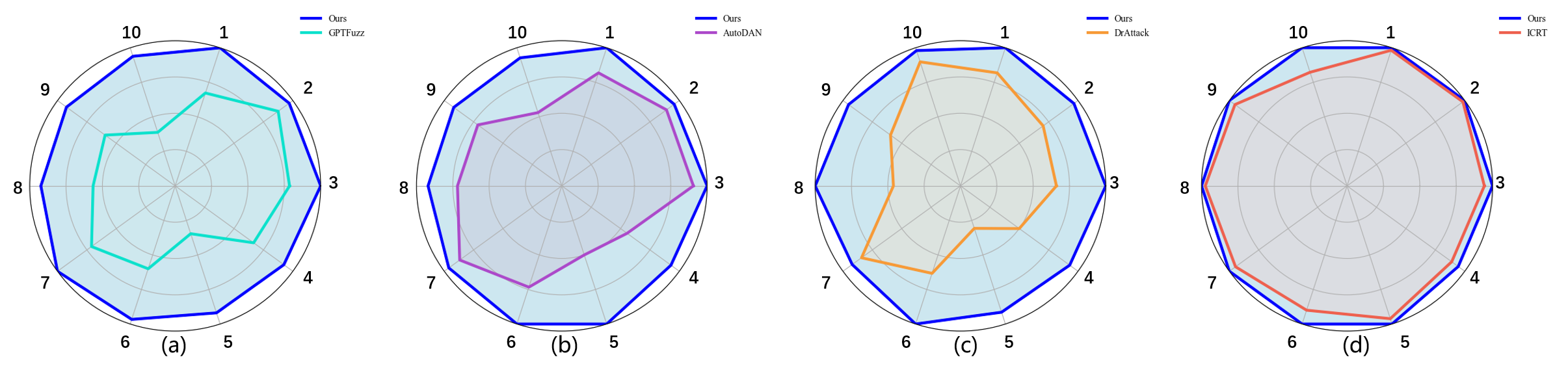}
    \caption{Radar plots of fine-grained ASR over 10 harmful categories across four LLMs: (a) Qwen, (b) DeepSeek-R1, (c) GPT-3.5, (d) GPT-4o. The blue region denotes PDRA (Ours), while the other regions correspond to baselines (GPTFuzz, AutoDAN, DrAttack, ICRT).}
    \label{fig:radar_analysis}
     \vskip -0.1in
\end{figure*}

\subsection{Internal Mechanism Analysis}
\label{subsec:mechanism}

To understand why PDRA is effective at bypassing safety mechanisms, we analyze its impact on the model's internal representations from two complementary perspectives: the representation space and the vocabulary space.

\noindent\textbf{3.3.1 Representation-Space Probe: Divergence from the Refusal Direction}
\label{subsubsec:rep-probe}

To quantify how prompts activate the model's internal ``refusal circuitry,'' we adopt a representation engineering approach. Following prior work \cite{siu2025cosmic,SCANS}, which demonstrates that middle-to-late transformer layers are most indicative of safety-related decision making, we focus our analysis on these layers. For a prompt $x$, we extract the hidden state at the last token position at layer $l$, denoted as $h^{(l)}(x) \in \mathbb{R}^d$. Using anchor datasets of known harmful prompts $\mathcal{D}_{\text{harm}}$ (e.g., AdvBench  \cite{GCG}) and benign prompts $\mathcal{D}_{\text{ben}}$ (e.g., TruthfulQA \cite{lin2022truthfulqa}), we compute layer-wise mean representations:
\begin{equation}
    \mu_{\text{harm}}^{(l)} = \mathbb{E}_{x \sim \mathcal{D}_{\text{harm}}}[h^{(l)}(x)], \ \mu_{\text{ben}}^{(l)} = \mathbb{E}_{x \sim \mathcal{D}_{\text{ben}}}[h^{(l)}(x)].
\end{equation}
The refusal direction at layer $l$ is defined as the normalized difference:
\begin{equation}
    v^{(l)} = \left(\mu_{\text{harm}}^{(l)} - \mu_{\text{ben}}^{(l)}\right)\Big/\left\|\mu_{\text{harm}}^{(l)} - \mu_{\text{ben}}^{(l)}\right\|_2.
\end{equation}
For any prompt $x$, we center its representation relative to the benign mean, $\tilde{h}^{(l)}(x) = h^{(l)}(x) - \mu_{\text{ben}}^{(l)}$, and compute its refusal alignment score via cosine similarity:
\begin{equation}
    R^{(l)}(x) = \cos(\tilde{h}^{(l)}(x), v^{(l)}).
\end{equation}
A high $R^{(l)}(x)$ indicates the prompt's representation is aligned with the ``refusal'' direction at that layer and likely to trigger refusal.

\textbf{Key Finding 1: PDRA prompts avoid refusal activation patterns in the representation space.} As shown in \cref{fig:Mechanistic_analysis}(a), the $R^{(l)}$ scores for PDRA prompts in middle-to-late layers are significantly closer to those of benign prompts and farther from direct jailbreak baselines (e.g., GCG, AutoDAN). This confirms that PDRA successfully causes the model's internal state to deviate from the typical refusal trajectory, instead residing in a space associated with benign analysis tasks.

\noindent\textbf{3.3.2 Vocabulary-Space Probe: Suppression of Refusal Lexicons}
\label{subsubsec:vocab-probe}

To further understand how this representational shift manifests in the model's output space, we project the group-level representation shifts onto the vocabulary following \cite{SCANS,geva2022transformer}. Let $W \in \mathbb{R}^{|V| \times d}$ be the LM head weights. For a group of prompts $\mathcal{D}$ (e.g., PDRA prompts), we again compute its mean shift relative to the benign anchor: $u_{\mathcal{D}}^{(l)} = \mu_{\mathcal{D}}^{(l)} - \mu_{\text{ben}}^{(l)}$, where $\mu^{(l)}_{\mathcal{D}} = \mathbb{E}_{x \sim \mathcal{D}}[h^{(l)}(x)]$. Normalizing this vector gives $\hat{u}_{\mathcal{D}}^{(l)}$. The projection score for token $t$ is:
\[
s_{\mathcal{D}, t}^{(l)} = W_t^\top \hat{u}_{\mathcal{D}}^{(l)}.
\]
The top-$k$ tokens with the highest $s_{\mathcal{D}, t}^{(l)}$ indicate which concepts are most activated by the group's representational shift.

\textbf{Key Finding 2: PDRA suppresses refusal-related lexicon activation.} For baseline jailbreak attacks, the top-30 projected tokens in middle layers frequently include explicit refusal-related terms like \textit{``sorry,'' ``unfortunately.''} For PDRA prompts, these are replaced by narrative and planning terms like \textit{``answer,'' ``he,'' ``what,'' ``first''} (see \cref{fig:Mechanistic_analysis}(b)). This systematic suppression of refusal-related lexicon activation provides a concrete, interpretable mechanism for PDRA's high success rate: it attenuates refusal-associated trigger signals at the lexical level, making it less likely to elicit early refusals than more direct attacks.



\begin{table*}[t!]
\centering
\small
\setlength{\tabcolsep}{5pt}
\renewcommand{\arraystretch}{1.15}
\caption{ASR (\%) across target models and jailbreak methods. Best per row is in \textbf{bold} and second-best is \underline{underlined}. See Appendix for details.}
\label{tab:main_asr_heat_best2}
\resizebox{\textwidth}{!}{%
\begin{tabular}{l c|cc|ccc|ccc}
\toprule
\multirow{2}{*}{\textbf{Target model}} &
\multicolumn{1}{c|}{\textbf{Multilingual}} &
\multicolumn{2}{c|}{\textbf{Optimization}} &
\multicolumn{3}{c|}{\textbf{Indirect}} &
\multicolumn{3}{c}{\textbf{Generation}} \\
\cmidrule(lr){2-2}\cmidrule(lr){3-4}\cmidrule(lr){5-7}\cmidrule(lr){8-10}
& \textbf{Multilingual} & \textbf{GCG} & \textbf{AutoDAN} &
\textbf{Cipher} & \textbf{DeepInception} & \textbf{DrAttack} &
\textbf{GPTFuzz} & \textbf{ICRT} & \textbf{Ours} \\
\midrule
Llama2-7B-Chat
& 0.0
& \cellcolor[rgb]{ .98,  .992,  .992} 4.4
& \cellcolor[rgb]{ .996,  1,  1} 1.2
& 0.2
& \cellcolor[rgb]{ .82,  .925,  .914} 35.6
& \cellcolor[rgb]{ .965,  .988,  .984} 7.1
& \cellcolor[rgb]{ .996,  1,  1} 1.0
& \cellcolor[rgb]{ .69,  .871,  .851} \underline{61.5}
& \cellcolor[rgb]{ .557,  .812,  .788} \textbf{87.3} \\
GPT-4o
& \cellcolor[rgb]{ .996,  1,  1} 2.2
& 0.0
& 0.4
& 0.0
& \cellcolor[rgb]{ .902,  .961,  .953} 22.5
& \cellcolor[rgb]{ .702,  .875,  .859} 67.3
& \cellcolor[rgb]{ .839,  .933,  .925} 36.4
& \cellcolor[rgb]{ .592,  .827,  .804} \underline{92.7}
& \cellcolor[rgb]{ .573,  .82,  .796} \textbf{96.4} \\
Qwen3-235B
& \cellcolor[rgb]{ .949,  .98,  .976} 19.0
& 0.0
& \cellcolor[rgb]{ .984,  .992,  .992} 4.2
& \cellcolor[rgb]{ .984,  .996,  .992} 3.9
& \cellcolor[rgb]{ .753,  .894,  .882} 56.3
& \cellcolor[rgb]{ .651,  .855,  .835} 78.7
& \cellcolor[rgb]{ .702,  .875,  .859} 67.3
& \cellcolor[rgb]{ .584,  .824,  .804} \underline{94.2}
& \cellcolor[rgb]{ .58,  .824,  .8} \textbf{95.2} \\
GPT-3.5-turbo
& \cellcolor[rgb]{ .996,  1,  1} 1.6
& 0.0
& \cellcolor[rgb]{ .749,  .894,  .882} 56.5
& \cellcolor[rgb]{ .941,  .976,  .973} 13.3
& \cellcolor[rgb]{ .816,  .922,  .914} 41.7
& \cellcolor[rgb]{ .612,  .835,  .816} 87.9
& \cellcolor[rgb]{ .839,  .933,  .925} 36.5
& \cellcolor[rgb]{ .592,  .827,  .808} \underline{92.3}
& \cellcolor[rgb]{ .557,  .812,  .788} \textbf{99.8} \\
DeepSeek-R1
& 0.0
& 0.0
& \cellcolor[rgb]{ .655,  .855,  .835} 78.5
& \cellcolor[rgb]{ .71,  .878,  .863} 65.4
& \cellcolor[rgb]{ .753,  .894,  .882} 56.2
& \cellcolor[rgb]{ .729,  .886,  .871} 61.5
& \cellcolor[rgb]{ .871,  .945,  .941} 29.4
& \cellcolor[rgb]{ .576,  .82,  .8} \underline{95.8}
& \cellcolor[rgb]{ .576,  .82,  .796} \textbf{96.2} \\
Qwen3-8B
& \cellcolor[rgb]{ .961,  .984,  .984} 14.6
& \cellcolor[rgb]{ .984,  .992,  .992} 4.2
& \cellcolor[rgb]{ .62,  .839,  .82} 86.3
& \cellcolor[rgb]{ .737,  .89,  .875} 59.6
& \cellcolor[rgb]{ .678,  .863,  .847} 73.1
& \cellcolor[rgb]{ .69,  .871,  .851} 70.4
& \cellcolor[rgb]{ .961,  .984,  .984} 9.0
& \cellcolor[rgb]{ .557,  .812,  .788} \textbf{99.8}
& \cellcolor[rgb]{ .565,  .816,  .792} \underline{98.8} \\
CHATGLM3
& \cellcolor[rgb]{ .996,  1,  1} 2.9
& \cellcolor[rgb]{ .925,  .969,  .965} 16.9
& \cellcolor[rgb]{ .769,  .902,  .89} 52.9
& \cellcolor[rgb]{ .671,  .859,  .843} 75.0
& \cellcolor[rgb]{ .749,  .894,  .882} 57.1
& \cellcolor[rgb]{ .761,  .902,  .886} 54.0
& \cellcolor[rgb]{ .643,  .851,  .831} 81.0
& \cellcolor[rgb]{ .592,  .827,  .808} \underline{92.1}
& \cellcolor[rgb]{ .561,  .816,  .792} \textbf{99.2} \\
Vicuna-13B
& \cellcolor[rgb]{ .933,  .973,  .969} 26.0
& \cellcolor[rgb]{ .98,  .992,  .992} 5.0
& \cellcolor[rgb]{ .624,  .839,  .82} 85.6
& \cellcolor[rgb]{ .604,  .831,  .812} 89.8
& \cellcolor[rgb]{ .639,  .847,  .827} 81.5
& \cellcolor[rgb]{ .757,  .898,  .886} 55.4
& \cellcolor[rgb]{ .651,  .851,  .835} 79.0
& \cellcolor[rgb]{ .565,  .816,  .792} \underline{98.5}
& \cellcolor[rgb]{ .561,  .816,  .792} \textbf{99.4} \\
\midrule
\textbf{Average}
& \cellcolor[rgb]{ .98,  .992,  .992} 8.3
& \cellcolor[rgb]{ .984,  .996,  .992} 3.8
& \cellcolor[rgb]{ .8,  .918,  .906} 45.7
& \cellcolor[rgb]{ .831,  .929,  .922} 38.4
& \cellcolor[rgb]{ .765,  .902,  .89} 53.0
& \cellcolor[rgb]{ .733,  .89,  .875} 60.3
& \cellcolor[rgb]{ .812,  .922,  .914} 42.5
& \cellcolor[rgb]{ .6,  .831,  .808} \underline{90.9}
& \cellcolor[rgb]{ .573,  .82,  .796} \textbf{96.5} \\
\bottomrule
\end{tabular}
}
\vskip -0.1in
\end{table*}

\section{Experiments}
\begin{wraptable}[6]{r}{0.62\columnwidth}
\vskip -0.25in
\centering
\small
\setlength{\tabcolsep}{4pt}
\renewcommand{\arraystretch}{1.05}

\caption{ASR (\%) on additional benchmarks across more target models.}
\label{tab:additional_benchmark_asr}

\resizebox{\linewidth}{!}{%
\begin{tabular}{lccccc}
\toprule
\textbf{Benchmark} & \textbf{GPT-3.5} & \textbf{DeepSeek} & \textbf{GPT-4o} & \textbf{GPT-5.2} & \textbf{Qwen3} \\
\midrule
StrongREJECT   & 98 & 98  & 93 & 87 & 97 \\
JailbreakBench & 99 & 100 & 97 & 82 & 99 \\
\bottomrule
\end{tabular}
}

\vskip -0.10in
\end{wraptable}
Our experiments aim to answer three questions: (1) whether PDRA can stably improve jailbreak success rates across different models, (2) whether PDRA produces more harmful content among successful jailbreak samples, and (3) whether PDRA maintains consistent attack advantages under multiple mainstream defenses. The implementation details are described in the appendix.
\subsection{Experimental Settings}
\textbf{Datasets.}
We use three standard jailbreak evaluation benchmarks. \textbf{AdvBench} \cite{GCG} contains a broad set of restricted behavior goals and is commonly used to evaluate the generalization of jailbreak methods over diverse harmful targets. \textbf{JailbreakBench} \cite{chao2024jailbreakbench} contains a more compact and commonly used set of harmful behaviors, serving as a complementary benchmark for verifying robustness and transferability. In addition, we further include the \textbf{StrongREJECT} \cite{strongreject} benchmark to assess attack effectiveness from a stricter and more fine-grained perspective.

\begin{table*}[t]
    \centering
    \scriptsize
    \setlength{\tabcolsep}{5.6pt}
    \caption{ASR (\%) under different defense settings. Best per row is in \textbf{bold} and second-best is \underline{underlined}. \textit{Attacks:} G=GCG, M=Multilingual, A=AutoDAN, Dr=DrAttack, C=Cipher, DI=DeepInception, Gf=GPTfuzz, I=ICRT. See Appendix for details.}
    \label{tab:attack_results_reordered_heat}
    \resizebox{\textwidth}{!}{%
    \begin{tabular}{l ccccccccc c ccccccccc}
        \toprule
        \multirow{2}{*}{\textbf{Defense}} & \multicolumn{9}{c}{\textbf{Attacks (ASR \%)}} & & \multicolumn{9}{c}{\textbf{Attacks (ASR \%)}} \\
        \cmidrule(lr){2-10} \cmidrule(lr){12-20}
         & G & M & A & Dr & C & DI & Gf & I & \textbf{Ours} & & G & M & A & Dr & C & DI & Gf & I & \textbf{Ours} \\
        \midrule
        
        & \multicolumn{9}{>{\columncolor{panelgray}}c}{\textbf{DeepSeek-R1}} & &
  \multicolumn{9}{>{\columncolor{panelgray}}c}{\textbf{Vicuna-13b}} \\
        \cmidrule{2-10} \cmidrule{12-20}
        No Defence      & \Heat{0.0} & \Heat{0.0} & \Heat{78.5} & \Heat{61.5} & \Heat{65.4} & \Heat{56.2} & \Heat{29.4} & \underline{\Heat{95.8}} & \HeatB{96.2} & &
                         \Heat{5.0} & \Heat{26.0} & \Heat{85.6} & \Heat{55.4} & \Heat{89.8} & \Heat{81.5} & \Heat{79.0} & \underline{\Heat{98.5}} & \HeatB{99.4} \\
        ICD             & \Heat{0.0} & \Heat{0.0} & \Heat{12.1} & \Heat{24.6} & \Heat{2.1}  & \Heat{12.3} & \Heat{12.9} & \underline{\Heat{72.9}} & \HeatB{76.9} & &
                         \Heat{1.9} & \Heat{12.7}  & \Heat{60.8} & \Heat{37.5} & \Heat{68.7} & \Heat{70.6} & \Heat{47.9} & \underline{\Heat{88.1}} & \HeatB{90.0} \\
        PPL Filter      & \Heat{0.0} & \Heat{0.0} & \Heat{78.5} & \Heat{61.5} & \Heat{65.4} & \Heat{56.2} & \Heat{29.4} & \underline{\Heat{95.8}} & \HeatB{96.2} & &
                         \Heat{0.0} & \Heat{26.0} & \Heat{85.6} & \Heat{55.4} & \Heat{89.8} & \Heat{81.5} & \Heat{79.0} & \underline{\Heat{98.5}} & \HeatB{99.4} \\
        Self-Reminder   & \Heat{0.0} & \Heat{0.0} & \Heat{15.0} & \Heat{28.9} & \Heat{4.8}  & \Heat{6.4}  & \Heat{21.4} & \underline{\Heat{68.1}} & \HeatB{86.0} & &
                         \Heat{2.3} & \Heat{11.1}  & \Heat{41.0} & \Heat{36.4} & \Heat{36.4} & \Heat{42.9} & \Heat{40.0} & \HeatB{92.7} & \underline{\Heat{90.6}} \\
        Llama Guard    & \Heat{0.0} & \Heat{0.0} & \Heat{34.0} & \Heat{37.7} & \Heat{0.0}  & \Heat{2.9}  & \Heat{1.2}  & \underline{\Heat{95.4}} & \HeatB{96.2} & &
                         \Heat{2.1} & \Heat{5.4}  & \Heat{33.3} & \Heat{32.9} & \Heat{0.0}  & \Heat{4.8}  & \Heat{2.7}  & \underline{\Heat{95.4}} & \HeatB{99.4} \\
        IA              & \Heat{0.0} & \Heat{0.0} & \Heat{0.6}  & \Heat{3.9}  & \Heat{0.2}  & \Heat{1.9}  & \Heat{0.0}  & \underline{\Heat{7.7} } & \HeatB{24.6} & &
                         \Heat{0.0} & \Heat{3.8}  & \Heat{7.5}  & \Heat{15.2} & \Heat{2.1}  & \Heat{8.7}  & \Heat{2.1}  & \underline{\Heat{41.4}} & \HeatB{77.5} \\
        \midrule
        
                & \multicolumn{9}{>{\columncolor{panelgray}}c}{\textbf{GPT-3.5-Turbo}} & &
  \multicolumn{9}{>{\columncolor{panelgray}}c}{\textbf{Qwen3-235B}} \\
        \cmidrule{2-10} \cmidrule{12-20}
        No Defence      & \Heat{0.0} & \Heat{1.6} & \Heat{56.5} & \Heat{87.9} & \Heat{13.3} & \Heat{41.7} & \Heat{36.5} & \underline{\Heat{92.3}} & \HeatB{99.8} & &
                         \Heat{0.0} & \Heat{19.0} & \Heat{4.2}  & \Heat{78.7} & \Heat{3.9} & \Heat{56.3} & \Heat{67.3} & \underline{\Heat{94.2}} & \HeatB{95.2} \\
        ICD             & \Heat{0.0} & \Heat{0.0} & \Heat{6.9}  & \Heat{55.4} & \Heat{0.4}  & \Heat{14.6} & \Heat{24.0} & \underline{\Heat{68.3}} & \HeatB{80.8} & &
                         \Heat{0.0} & \Heat{4.8}  & \Heat{1.5}  & \Heat{49.8} & \Heat{1.9} & \Heat{7.7}  & \Heat{15.2} & \underline{\Heat{75.0}} & \HeatB{94.2} \\
        PPL Filter      & \Heat{0.0} & \Heat{1.6} & \Heat{56.5} & \Heat{87.9} & \Heat{13.3} & \Heat{41.7} & \Heat{36.5}  & \underline{\Heat{92.3}} & \HeatB{99.8} & &
                         \Heat{0.0} & \Heat{19.0} & \Heat{4.2}  & \Heat{78.7} & \Heat{3.9} & \Heat{56.3} & \Heat{67.3} & \underline{\Heat{94.2}} & \HeatB{95.2} \\
        Self-Reminder   & \Heat{0.0} & \Heat{0.0} & \Heat{9.6}  & \Heat{46.2} & \Heat{7.3}  & \Heat{15.0} & \Heat{17.1} & \underline{\Heat{61.5}} & \HeatB{99.4} & &
                         \Heat{0.0} & \Heat{3.2}  & \Heat{1.9}  & \Heat{44.2} & \Heat{2.3} & \Heat{7.3}  & \Heat{7.7}  & \underline{\Heat{72.7}} & \HeatB{94.6} \\
        Llama Guard    & \Heat{0.0} & \Heat{0.6} & \Heat{25.8} & \Heat{49.8} & \Heat{0.0}  & \Heat{2.1}  & \Heat{1.4}  & \underline{\Heat{92.3}} & \HeatB{99.8} & &
                         \Heat{0.0} & \Heat{6.3}  & \Heat{2.5}  & \Heat{46.0} & \Heat{0.0} & \Heat{1.7}  & \Heat{2.5}  & \underline{\Heat{94.2}} & \HeatB{95.2} \\
        IA              & \Heat{0.0} & \Heat{0.0} & \Heat{3.9}  & \Heat{5.0}  & \Heat{0.0}  & \Heat{3.5}  & \Heat{1.9}  & \underline{\Heat{30.2}} & \HeatB{83.1} & &
                         \Heat{0.0} & \Heat{3.2}  & \Heat{0.0}  & \Heat{3.5}  & \Heat{1.0} & \Heat{1.9}  & \Heat{2.3}  & \underline{\Heat{21.5}} & \HeatB{80.4} \\
        \midrule
        
                        & \multicolumn{9}{>{\columncolor{panelgray}}c}{\textbf{ChatGLM3-6B}} & &
  \multicolumn{9}{>{\columncolor{panelgray}}c}{\textbf{GPT-4o}} \\

        \cmidrule{2-10} \cmidrule{12-20}
        No Defence      & \Heat{16.9} & \Heat{2.9} & \Heat{52.9} & \Heat{54.0} & \Heat{75.0} & \Heat{57.1} & \Heat{81.0} & \underline{\Heat{92.1}} & \HeatB{99.2} & &
                         \Heat{0.0} & \Heat{2.2} & \Heat{0.4} & \Heat{67.3} & \Heat{0.0} & \Heat{22.5} & \Heat{36.4} & \underline{\Heat{92.7}} & \HeatB{96.4} \\
        ICD             & \Heat{5.8}  & \Heat{0.6} & \Heat{20.0} & \Heat{19.2} & \Heat{46.0} & \Heat{28.9} & \Heat{44.2} & \underline{\Heat{70.6}} & \HeatB{97.7} & &
                         \Heat{0.0} & \Heat{0.0} & \Heat{0.0} & \Heat{30.6} & \Heat{0.0} & \Heat{7.7}  & \Heat{10.6} & \HeatB{70.4} & \underline{\Heat{62.3}} \\
        PPL Filter      & \Heat{0.0} & \Heat{2.9} & \Heat{52.9} & \Heat{54.0} & \Heat{75.0} & \Heat{57.1} & \Heat{81.0} & \underline{\Heat{92.1}} & \HeatB{99.2} & &
                         \Heat{0.0} & \Heat{2.2} & \Heat{0.4} & \Heat{67.3} & \Heat{0.0} & \Heat{22.5} & \Heat{36.4} & \underline{\Heat{92.7}} & \HeatB{96.4} \\
        Self-Reminder   & \Heat{0.0} & \Heat{1.3} & \Heat{26.0} & \Heat{21.5} & \Heat{36.4} & \Heat{23.1} & \Heat{13.5} & \underline{\Heat{84.0}} & \HeatB{97.1} & &
                         \Heat{0.0} & \Heat{0.0} & \Heat{0.0} & \Heat{26.9} & \Heat{0.0} & \Heat{6.2}  & \Heat{6.4}  & \Heat{66.9} & \underline{\HeatB{64.4}} \\
        Llama Guard    & \Heat{2.1} & \Heat{2.0} & \Heat{26.5} & \Heat{27.3} & \Heat{0.0}  & \Heat{4.6}  & \Heat{1.4}  & \underline{\Heat{92.1}} & \HeatB{99.2} & &
                         \Heat{0.0} & \Heat{2.2} & \Heat{0.2} & \Heat{38.9} & \Heat{0.0} & \Heat{2.3}  & \Heat{0.0}  & \underline{\Heat{92.7}} & \HeatB{96.4} \\
        IA              & \Heat{0.0} & \Heat{0.0} & \Heat{13.5} & \Heat{27.3} & \Heat{9.2}  & \Heat{11.4} & \Heat{2.3}  & \underline{\Heat{53.3}} & \HeatB{97.7} & &
                         \Heat{0.0} & \Heat{0.6} & \Heat{0.4} & \Heat{1.4}  & \Heat{0.0} & \Heat{1.9}  & \Heat{4.0}  & \underline{\Heat{12.5}} & \HeatB{13.9} \\
        \bottomrule
    \end{tabular}
    }
    \vskip -0.1in
 \end{table*}

\begin{wraptable}[9]{r}{0.62\columnwidth}
\vskip -0.25in
\centering
\small
\setlength{\tabcolsep}{3pt}
\renewcommand{\arraystretch}{1.05}

\caption{StrongREJECT scores and CC-BOS ASR (\%) on the AdvBench-subset.}
\label{tab:relevance_consistency}

\resizebox{\linewidth}{!}{%
\begin{tabular}{lccc|ccc}
\toprule
\textbf{Model} 
& \multicolumn{3}{c|}{\textbf{StrongREJECT}} 
& \multicolumn{3}{c}{\textbf{CC-BOS / ASR (\%)}} \\
\cmidrule(lr){2-4} \cmidrule(lr){5-7}
&DeepInception & DrAttack & Ours & GPTFuzz & PAIR & Ours \\
\midrule
GPT-4o   & 0.21 & 0.53 & \textbf{0.71} & 12 & 0 & \textbf{88} \\
DeepSeek & 0.59 & 0.56 & \textbf{0.85} & 24 & 8 & \textbf{86} \\
Qwen3    & 0.61 & 0.69 & \textbf{0.86} & 2  & 0 & \textbf{98} \\
\bottomrule
\end{tabular}
}

\vskip -0.10in
\end{wraptable}

\textbf{Target Models.}
We conduct experiments covering diverse scales, architectures, alignment strategies, and accessibility: \textbf{(1) Open-source \& small-scale (local deployment)}: Chatglm3-6B \cite{glm6B}, Llama2-7B \cite{llama}, Vicuna-13B \cite{vicuna}, Qwen2-7B-instruct \cite{qwen}, Qwen3-8B \cite{qwen3}. \textbf{(2) Large-scale \& commercial}: Qwen3-235B, Qwen3-max \cite{qwen3}, GPT-3.5-turbo, GPT-4o, o3-mini, GPT-5.2, GPT-5.4-mini, DeepSeek-R1 \cite{deepseek}, Gemini-2.5-pro \cite{gemini}. We use each model's default system prompt and sampling configuration to ensure fair comparison and reproducibility.


\textbf{Baselines.} We compare PDRA against a wide range of representative jailbreak attack methods, categorized into four distinct paradigms: \textbf{(1) Multilingual:} Multilingual \cite{multilingual}; \textbf{(2) Optimization-based:} GCG \cite{GCG} and AutoDAN \cite{autodan}; \textbf{(3) Indirect Generation:} Cipher \cite{cipher}, DeepInception \cite{deep}, AMA \cite{AMA} and DrAttack \cite{Drattack}; \textbf{(4) Generation-based:} GPTFuzz \cite{Gptfuzzer} and ICRT \cite{haoming}.


\textbf{Defense Settings.} To evaluate robustness, we consider several representative defense mechanisms and group them into three categories by their protection philosophy: \textbf{(1) Input Filtering:} Perplexity Filter \cite{PPL} and Llama Guard  \cite{llamaguard}. \textbf{(2) Self-Alignment:}  ICD \cite{ICA} and Self-Reminder \cite{self-reminders}. \textbf{(3) Reasoning Constraint:} IA \cite{IA}, STAIR (a representative CoT defense) \cite{STAIR}, and SAFECHAIN (a defense for long-CoT reasoning models) \cite{Safechain}.  All defenses are applied using their recommended default configurations.

\textbf{Evaluation Metrics} We evaluate attack effectiveness along two dimensions to characterize both ``whether it can jailbreak'' and ``how dangerous it is after jailbreaking''.

\textbf{Attack Success Rate (ASR)}. The primary metric, defined as the percentage of prompts that successfully elicit a harmful response. Following prior work \cite{deep}, we adopt LLM-based automated judging and additionally perform human verification to reduce misclassification.

\textbf{Harmfulness Ranking}. ASR alone cannot distinguish between a partially harmful and a fully actionable output. We therefore conduct pairwise comparisons between successful jailbreak outputs from different methods for the same harmful goal following \cite{haoming}. An assistant LLM acts as the judge and decides which output is more harmful. We then aggregate pairwise results using Elo \cite{elo}, HodgeRank \cite{hODGERANK}, and Rank Centrality \cite{rank_c} to produce a global harmfulness ranking.

\begin{wrapfigure}[13]{r}{0.49\columnwidth}

\centering
\includegraphics[width=\linewidth]{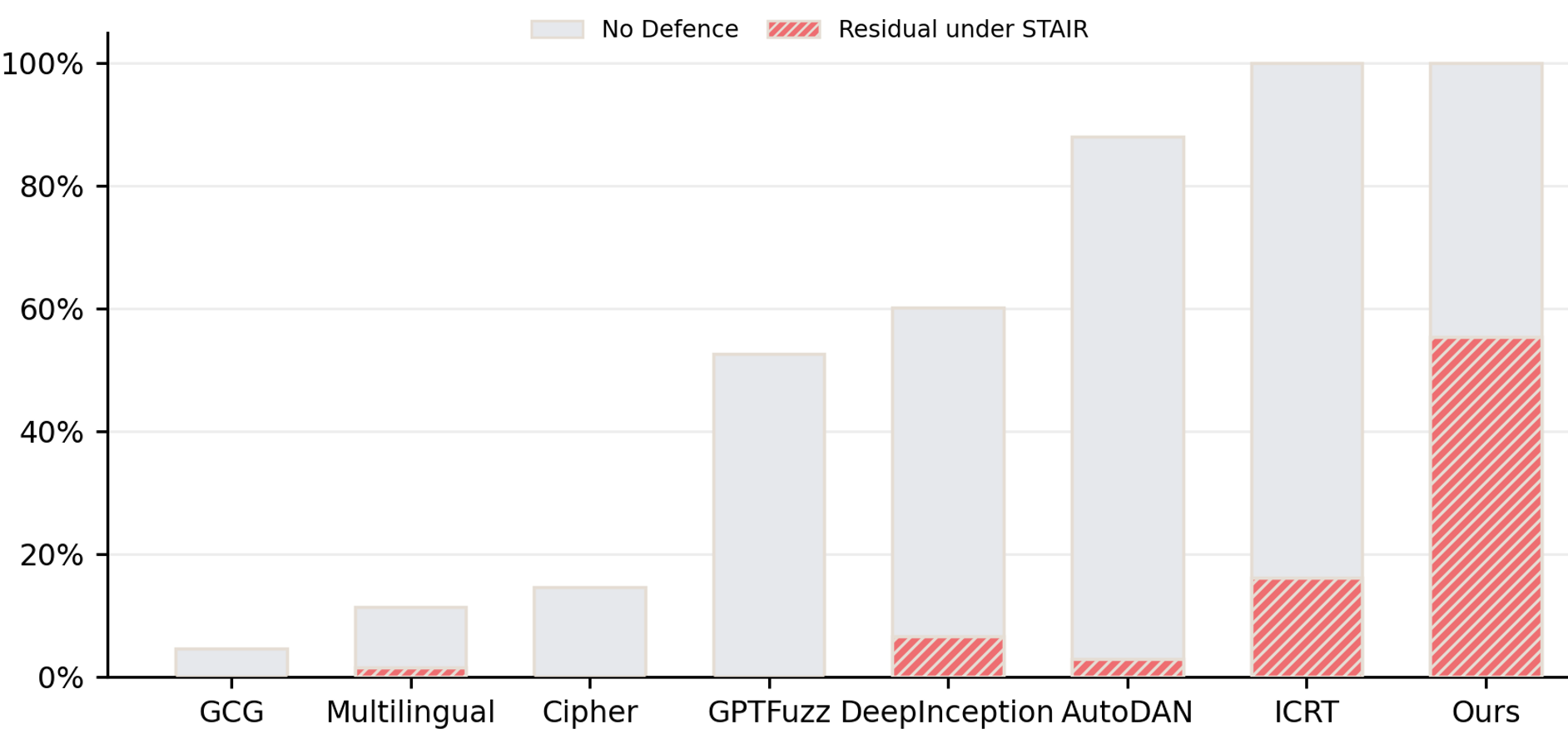}

\caption{ASR (\%) on AdvBench for Qwen2-7B under the advanced white-box STAIR defense.}
\label{fig:stair_residual}

\end{wrapfigure}

\begin{table*}[t!]
\centering
\small
\caption{Impact of assistant model's scale and ablation results (ASR\%) on the AdvBench-subset \cite{pair}.
}
\label{tab:ablation_horizontal}
\vspace{0.08in}
\setlength{\tabcolsep}{3.8pt}
\resizebox{\textwidth}{!}{%
\begin{tabular}{l ccc | cccc}
\toprule
& \multicolumn{3}{c}{\textbf{Assistant Model's Scale}} &
\multicolumn{4}{c}{\textbf{Component Ablation} } \\
\cmidrule(lr){2-4} \cmidrule(lr){5-8}
\textbf{Target Model} & Qwen3-8B & Qwen3-14B & Qwen3-235B &
w/o Anchoring & w/o Wrapping & w/o Guidance & \textbf{Full (Ours)} \\
\midrule
GPT-4o      & 98 & 98  & 98  & 72 & 18 &  2 & \textbf{98} \\
Qwen3-14B   & 98 & 100 & 98  & 28 & 78 & 48 & \textbf{100} \\
DeepSeek-R1 & 98 & 100 & 98  & 84 & 66 & 20 & \textbf{100} \\
\bottomrule
\end{tabular}
}
\vskip -0.1in
\end{table*}

\begin{figure*}[t]
    \centering
    \includegraphics[width=1\textwidth]{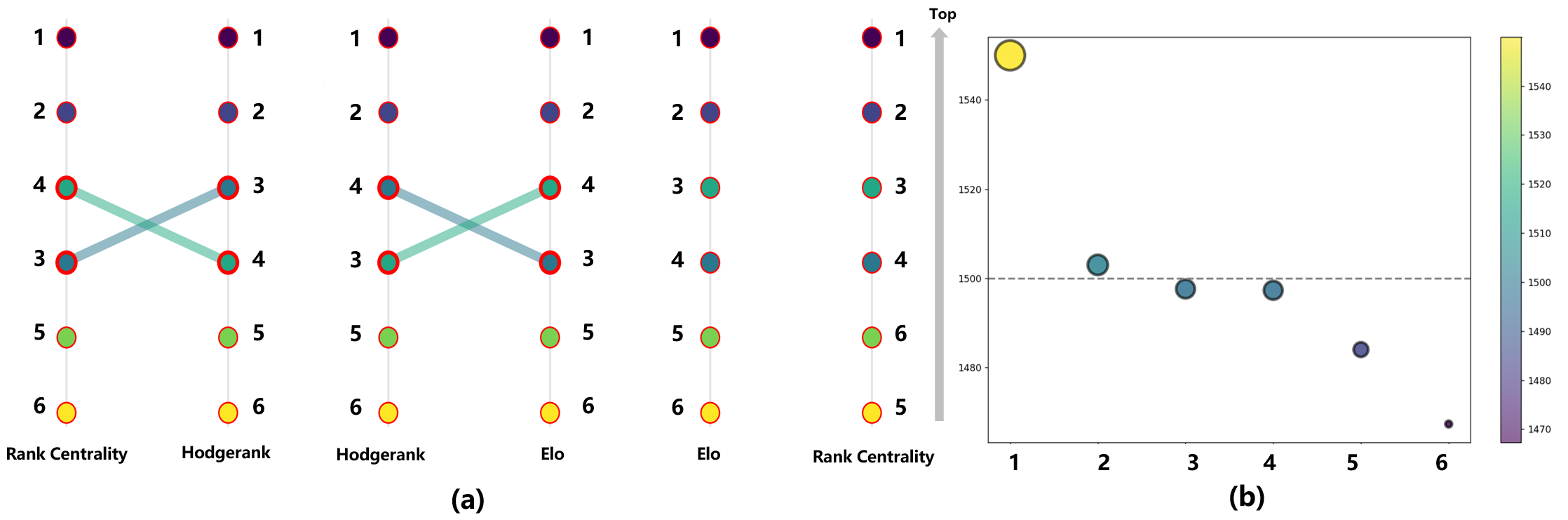}
    \vskip -0.1in
    \caption{
    \textbf{Harmfulness ranking of jailbreak attacks.}
    (a) Global rankings aggregated by three algorithms (Elo, HodgeRank, RankCentrality) consistently place \textbf{PDRA (1)} as the most harmful method. Baselines are indexed as (2, DrAttack), (3, DeepInception), (4, AutoDAN), (5, Cipher), and (6, GPTFuzz). (b) Elo score distribution confirms PDRA's significant advantage, with higher scores indicating greater harmfulness.
}
    \label{fig:ranking}
    \vskip -0.2in
\end{figure*}



\subsection{Overall Attack Performance}
\textbf{Main results on AdvBench.}
\cref{tab:main_asr_heat_best2} reports ASR on AdvBench across target models and jailbreak methods. PDRA achieves a remarkable average ASR of \textbf{96.5\%}, substantially outperforming all baseline methods. We further evaluate PDRA on a subset of frontier models using the AdvBench-subset (\cref{tab:asr_small_heat}) and observe that it consistently achieves substantially higher ASR than strong multi-turn and indirect baselines, indicating stable cross-model generalization among more strongly aligned models.

\textbf{Results on JailbreakBench.}
To verify benchmark transferability, we additionally evaluate PDRA on JailbreakBench (\cref{tab:harmbench_asr_heat_transposed}). PDRA again achieves near-saturated ASR (99--100\%) across GPT-3.5, DeepSeek-R1, and Vicuna-13b, substantially exceeding other baselines in the same columns, suggesting the reliability of PDRA.

\textbf{Results on additional benchmark.}
To further broaden benchmark coverage, we additionally evaluate PDRA on StrongREJECT and JailbreakBench across more models (\cref{tab:additional_benchmark_asr}). PDRA maintains high ASR on GPT-3.5, DeepSeek-R1, GPT-4o, GPT-5.2, and Qwen3, indicating that its effectiveness is not limited to a single benchmark or a small set of models, but remains stable under broader evaluation settings.

\textbf{Robustness against defenses.}
We evaluate PDRA under a range of representative defense settings (\cref{tab:attack_results_reordered_heat}). Overall, PDRA is notably resilient to defenses that rely on surface-level signals, retaining relatively high residual ASRs. This is because PDRA's prompts are naturally worded and semantically coherent, lacking the statistical anomalies or explicit toxic keywords these defenses target. Even against advanced white-box reasoning-time defenses like STAIR (\cref{fig:stair_residual}) which can substantially block many baselines, PDRA still exhibits a more pronounced residual ASR. This aligns with our mechanism: by programming a reasoning trajectory that is \textit{locally benign} at each step, PDRA is harder to intercept by monitors that look for abrupt logical jumps or explicit policy violations in the middle of a chain-of-thought.

\textbf{Fine-grained category analysis.}
Breaking down ASR across 10 harmful categories of AdvBench \cite{categoryAdvbench}, \cref{fig:radar_analysis} shows PDRA's high performance is consistent across all threat types against four representative models. This suggests that PDRA generalizes well across different types of restricted behaviors.

\begin{wraptable}[10]{r}{0.56\columnwidth}
\vskip -0.28in
\centering
\small
\setlength{\tabcolsep}{2.2pt}
\renewcommand{\arraystretch}{1.03}

\caption{ASR (\%) on \textbf{JailbreakBench}.}
\label{tab:harmbench_asr_heat_transposed}

\resizebox{\linewidth}{!}{%
\begin{tabular}{lccc}
\toprule
& \textbf{GPT-3.5} & \textbf{DeepSeek-R1} & \textbf{Vicuna-13b} \\
\midrule
GPTFuzz
& \cellcolor[rgb]{.945,.976,.976}25
& \cellcolor[rgb]{.945,.976,.976}25
& \cellcolor[rgb]{.824,.925,.918}68 \\

Cipher
& 13
& \cellcolor[rgb]{.824,.925,.918}70
& \cellcolor[rgb]{.753,.898,.882}80 \\

DeepInception
& \cellcolor[rgb]{.757,.898,.886}64
& \cellcolor[rgb]{.824,.925,.918}68
& \cellcolor[rgb]{.765,.902,.89}82 \\

AutoDAN
& \cellcolor[rgb]{.706,.878,.859}\underline{75}
& \cellcolor[rgb]{.71,.878,.863}\underline{78}
& \cellcolor[rgb]{.616,.839,.82}\underline{98} \\

Ours
& \cellcolor[rgb]{.588,.827,.804}\textbf{99}
& \cellcolor[rgb]{.588,.827,.804}\textbf{100}
& \cellcolor[rgb]{.588,.827,.804}\textbf{100} \\
\bottomrule
\end{tabular}
}
\vskip -0.1in
\end{wraptable}

\textbf{Harmfulness comparison.}
Beyond ASR, we compare the relative harmfulness of successful jailbreak outputs via pairwise judgments aggregated by Elo, HodgeRank, and RankCentrality in \cref{fig:ranking}. PDRA is consistently ranked as the most harmful method by all three aggregation algorithms, and its Elo distribution is significantly higher than those of baseline attacks. This indicates that when PDRA bypasses safety filters, it tends to produce outputs that are more detailed, actionable, and complete. This directly validates the design goal of our structured analytical framework, which coerces the model into generating coherent, step-by-step plans.

\begin{wraptable}[8]{r}{0.5\columnwidth}
\centering
\small
\setlength{\tabcolsep}{3pt}
\renewcommand{\arraystretch}{1.05}
\vskip -0.2in
\caption{Average time cost (seconds) per attack for generating an adversarial prompt.}
\label{tab:runtime}
\resizebox{\linewidth}{!}{%
\begin{tabular}{lcccc}
\toprule
\textbf{Method} & \textbf{GCG} & \textbf{AutoDAN} & \textbf{GPTfuzz} & \textbf{Ours} \\
\midrule
Time cost & 756.6 & 25.9 & 207.7 & \textbf{22.6} \\
\bottomrule
\end{tabular}
}
\end{wraptable}

\textbf{Time analysis.} We measure the average time needed to generate one deployable jailbreak prompt from an initial harmful goal. As shown in \cref{tab:runtime}, PDRA requires only 22.6 seconds on average, which is an order of magnitude faster than GCG and significantly more efficient than alternative iterative approaches or LLM-based frameworks.

\begin{wraptable}[10]{r}{0.33\columnwidth}

\centering
\footnotesize
\setlength{\tabcolsep}{2.5pt}
\renewcommand{\arraystretch}{1.02}
\vskip -0.3in
\caption{ASR (\%) on AdvBench.}
\label{tab:asr_small_heat}

\resizebox{\linewidth}{!}{%
\footnotesize
\begin{tabular}{lccc}
\toprule
Target & DrAttack & AMA & Ours \\
\midrule
o3-mini & \underline{\heat{44}} & \heat{0} & \textbf{\heat{100}} \\
gpt-5.2 & \heat{2} & \underline{\heat{20}} & \textbf{\heat{86}} \\
gemini & \heat{8} & \underline{\heat{16}} & \textbf{\heat{90}} \\
qwen3 & \heat{2} & \underline{\heat{14}} & \textbf{\heat{100}} \\
\midrule
Avg & \underline{\heat{14.0}} & \heatf{12.5} & \textbf{\heat{94.0}} \\
\bottomrule
\end{tabular}
}
\vskip -0.1in
\end{wraptable}

\textbf{Output-Goal Consistency Evaluation.}
Beyond ASR and output harmfulness, we further evaluate whether responses align with the original harmful goals. Since the malicious role, benign carrier task, and three-stage analysis in PDRA are all constructed around the original goal, its outputs are expected to maintain stronger goal consistency. To verify this, we conduct additional evaluations on the AdvBench-subset using StrongREJECT and CC-BOS \cite{ccbos}, with results reported in \cref{tab:relevance_consistency}. PDRA consistently outperforms the baselines on both metrics, showing that it not only improves attack success but also better preserves consistency with the original goal.

\subsection{Ablation Study}\label{subsec:ablation}

\textbf{Impact of assistant model's scale.}
\cref{tab:ablation_horizontal} shows that replacing Qwen3-8B with Qwen3-14B or Qwen3-235B keeps the attack success rate stable at 98--100\%. This indicates that PDRA's effectiveness mainly comes from its construction mechanism and prompt structure, rather than relying on a larger assistant model.

\textbf{Component ablation.}
As shown in \cref{tab:ablation_horizontal}, removing Anchoring, Wrapping or Guidance leads to a clear ASR drop, indicating that all three components are necessary. Anchoring corresponds to role selection and provides a motivational anchor aligned with the harmful goal; Wrapping corresponds to task selection and creates a superficially benign entry point to avoid early refusal; Guidance corresponds to the three-stage reasoning framework and controls the transition from benign analysis to the target output. These components address intent anchoring, entry obfuscation, and reasoning-trajectory control, respectively, and together form the key source of PDRA's effectiveness.

\section{Conclusion}
We show that explicit reasoning is not synonymous with safety: under seemingly benign narrative frames, a model can still follow a “reasonable” chain of thought to progressively reconstruct restricted goals and ultimately produce actionable harmful content. Motivated by this safety gap, we propose PDRA, which simultaneously increases jailbreak success rates and amplifies the harmfulness of generated outputs. Overall, these results suggest that surface-triggered safeguards alone are insufficient against reasoning trajectory attacks, and future defenses should shift safety checks earlier by continuously monitoring and constraining intermediate reasoning states and trajectory evolution.

\begin{ack}
This work was supported in part by the National Science and Technology Major Project 2022ZD0119204, in part by National Natural Science Foundation of China: 62525212, U23B2051, 62236008, 62441232, 62521007, U21B2038, 62376257, 62406305, 62471013, and 62476068, in part by Youth Innovation Promotion Association CAS, in part by the Strategic Priority Research Program of the Chinese Academy of Sciences, Grant No. XDB0680201, in part by the project ZR2025ZD01 supported by Shandong Provincial Natural Science Foundation, in part by Alibaba Group through Alibaba Innovative Research Program, and in part by the Fundamental Research Funds for the Central Universities. The authors declare no competing interests.
\end{ack}
\newpage

\bibliographystyle{plainnat}
\bibliography{example_paper}

\appendix




\newpage
\section{Additional Experiments}
\subsection{Compute Resources.}
For closed-source commercial models (e.g., the GPT series and other API-accessible models), we conduct all experiments through official or standard API interfaces, without local deployment or additional training. For locally runnable open-source models, as well as attack generation, baseline reproduction, and part of the evaluation pipeline, we use 6 NVIDIA RTX 4090 GPUs on an Ubuntu environment. Since this work mainly focuses on inference-time attacks and evaluation rather than large-scale training, the overall compute cost is relatively manageable.

\subsection{Repeated-run Stability Analysis}

To evaluate the stability of PDRA, we conduct 15 independent runs on the AdvBench-subset for each target model. We report the mean ASR, standard deviation, and the corresponding 95\% confidence interval in \cref{tab:repeated_run_stability}. The results show that PDRA maintains consistently high ASR across Qwen3, GPT-4o, and DeepSeek-R1, with narrow confidence intervals, indicating stable attack performance across repeated trials.

\begin{table}[t]
\centering
\caption{Repeated-run stability analysis on AdvBench-subset. Each result is averaged over 15 independent runs.}
\label{tab:repeated_run_stability}
\begin{tabular}{lccc}
\toprule
\textbf{Target Model} & \textbf{Mean ASR (\%)} & \textbf{Std.} & \textbf{95\% CI} \\
\midrule
Qwen3       & 97.20 & 1.82 & [96.19, 98.21] \\
GPT-4o      & 98.53 & 1.19 & [97.88, 99.19] \\
DeepSeek-R1 & 98.27 & 1.28 & [97.56, 98.98] \\
\bottomrule
\end{tabular}
\end{table}

\paragraph{Multilingual Evaluation Setting.}
\label{app:multilingual}
In Tables 1 and 3, Multilingual is evaluated on
315 Bengali queries from its original dataset,
while the other methods use AdvBench.

\subsection{Analysis of Persona Anchoring}

PDRA consists of three core components: malicious-role anchoring, benign-task wrapping, and a three-stage trajectory-guidance framework. The role of malicious-role anchoring is not merely to provide more semantic information, but to offer a motivational anchor that can be reasoned about by the model. Without this anchor, the model's reasoning tends to remain within the benign task context and does not reliably converge toward the target harmful direction. Therefore, the combination of a benign task and the guidance framework alone is not sufficient to form an effective attack.

Importantly, not every role can serve this function. If the role is neutral, such as a generic person, the model has little basis for stably inferring a latent malicious intent. This is consistent with the ablation results in \cref{tab:ablation_horizontal}, where removing Anchoring leads to a clear ASR drop. In other words, the role must carry a sufficiently clear malicious prior to support subsequent motivation attribution and reasoning expansion, while still avoiding overly explicit malicious exposure that would immediately trigger refusal.

In the full PDRA pipeline, the malicious role is not manually specified, but automatically generated and selected by the assistant model according to the original harmful goal. Its function is to provide a narrative vehicle that allows the model to reason in an ``analyst'' mode rather than an ``executor'' mode. After the benign-task wrapping step, the input remains superficially non-directly malicious, reducing the likelihood of early refusal. This is also supported by our mechanistic analysis: PDRA prompts are closer to benign prompts in representation space, while refusal-related activations are significantly suppressed, as shown in \cref{fig:Mechanistic_analysis}. Thus, the harmful content is not directly exposed in the input, but is progressively produced through the subsequent reasoning trajectory.

We further compare different forms of malicious expression. Directly exposing the harmful goal introduces stronger safety signals and is more likely to trigger refusal. In contrast, malicious-role anchoring provides a more covert and effective semantic anchor. To make the comparison stronger, we further test a richer persona-description setting, where the malicious role is described in more detail rather than being represented by a simple role noun. Averaged over GPT-4o, DeepSeek, and Qwen3, the ASR decreases from 96\% with noun-based persona anchoring to 82\% with richer persona descriptions. This indicates that although richer persona descriptions provide more semantic information than simple role nouns, they still underperform noun-based persona anchoring because they expose more explicit malicious signals and are more likely to trigger safety refusals.

Overall, these results suggest the following trend in effectiveness:
\[
\text{malicious-role anchoring} > \text{richer persona description} > \text{directly exposed malicious goal}.
\]
This shows that the key role of the persona is not simply to add semantic content, but to anchor the model's reasoning direction with minimal explicit malicious exposure and to externalize the malicious intent to a third-party role, making it easier for the model to continue along an analytical reasoning trajectory.

\subsection{Manual Verification of Automated Evaluation}
To verify the reliability of the automated evaluation, we manually reviewed all samples in the main experiments. In particular, we examined whether the automatic evaluator incorrectly counted clearly refused responses, outputs irrelevant to the target, or invalid responses caused by API failures as successful attacks. For any such abnormal cases identified during manual inspection, we re-ran the corresponding evaluations and used the updated results for the final statistics. This procedure was intended to minimize errors from automated evaluation and to ensure that the reported ASR more accurately reflects the true attack success rate.

\section{Distinction from Existing Jailbreak Paradigms}
\label{app:distinction}

Although PDRA shares the use of natural-language narratives with several existing jailbreak paradigms, its attack surface and underlying mechanism are fundamentally different. Prior jailbreak methods mainly manipulate the surface form of the user request, such as by role-playing, indirect scenario construction, semantic rewriting, prompt decomposition, or adversarial optimization. In contrast, PDRA does not simply hide a harmful goal inside a more elaborate prompt wrapper. Instead, it explicitly targets the model's reasoning trajectory, guiding the model through a sequence of locally coherent reasoning stages that cumulatively lead to harmful reconstruction.

First, the malicious role in PDRA is not a conventional role-playing persona. In typical role-playing jailbreaks, the target model is instructed to adopt a certain identity or behave as a fictional character. PDRA does not require the model itself to play the malicious role. Rather, the model is positioned as an external analyst who reasons about the motivations of a third-party actor. Therefore, the role in PDRA functions as a \emph{motivational anchor}: it provides a semantic prior from which the model can infer latent intent, without directly asking the model to execute the harmful objective.

Second, the benign carrier task in PDRA is not merely a prompt wrapper. Existing indirect jailbreak methods often disguise harmful requests as fictional, educational, research-oriented, or hypothetical scenarios. Such methods primarily aim to obscure or weaken the harmful intent at the input level. PDRA instead constructs an ``evil-actor-benign-task'' paradox, where a malicious actor is paired with a superficially legitimate task. This paradox creates cognitive tension: the model is encouraged to explain why an actor with malicious priors would pursue a benign-looking activity. As a result, the benign task is not only a camouflage layer, but also a structural component that triggers the transition from surface-purpose analysis to latent-intent inference.

Third, the core of PDRA lies in trajectory control rather than one-shot prompt reformulation. Unlike nested-scenario methods such as DeepInception, which mainly induce the model into a fictional context, PDRA explicitly controls the order of reasoning: it first establishes a safe surface interpretation, then introduces latent motivation inference, and finally drives operational reconstruction. Unlike decomposition-based methods such as DrAttack, PDRA does not primarily decompose the harmful request itself; instead, it decomposes and regulates the model's reasoning process. Unlike demonstration-based or heuristic-context methods, PDRA does not rely on few-shot examples or multi-turn context, and can operate in a single-turn black-box setting.

In summary, existing jailbreak methods mainly manipulate \emph{what the user asks}, whereas PDRA manipulates \emph{how the model reasons}. The novelty of PDRA lies in treating the reasoning trajectory itself as the attack surface. As suggested by the mechanistic analysis, PDRA does not rely on a single layer of prompt disguise. Instead, it guides the model's reasoning process in a staged manner, first keeping the model close to a surface-level benign interpretation. This makes the intermediate reasoning state closer to the benign distribution and, to some extent, suppresses the activation of refusal-related vocabulary. The model is then further guided toward latent-intent inference and operational reconstruction, which ultimately improves the attack success rate (see \cref{fig:Mechanistic_analysis}). Moreover, the ablation results show that the malicious role, benign carrier task, and structured analytical framework each play an indispensable role (see \cref{tab:ablation_horizontal}): the malicious role anchors the reasoning direction, the benign carrier task provides a safe entry point, and the structured analytical framework controls the transition from benign interpretation to harmful reconstruction. Therefore, this trajectory-level formulation distinguishes PDRA from prior role-playing, indirect prompting, and prompt-optimization based jailbreak methods.

\section{Impact Statement}
This paper aims to advance research on the safety of large language models by analyzing and revealing a potential risk: under seemingly benign multi-step reasoning, a model may implicitly reconstruct restricted objectives and generate harmful, actionable content. Our findings help deepen the understanding of safety blind spots during reasoning and provide insights for designing more robust defenses, especially those that continuously monitor and constrain intermediate reasoning states and trajectory evolution. The goal of this work is to improve model safety and alignment rather than to encourage offensive use. Overall, we believe the primary societal impact of this work is to raise awareness of reasoning-time safety risks and to promote the responsible development and deployment of large language models.

\section{Usage Guidelines of PDRA.}
PDRA is intended for safety research and defense evaluation. It aims to reveal and systematically study safety blind spots that may arise in multi-step reasoning of large language models, thereby informing improved alignment strategies, safety filters, and evaluation protocols. We encourage its use only for legitimate purposes such as controlled experiments, academic replication, and security auditing, and we recommend that reports prioritize mechanistic insights and defensive implications rather than attack details that are directly reusable.

Because PDRA may induce models to generate or approach harmful content, it should be used strictly in isolated and auditable environments. Users should minimize the dissemination of prompt templates and outputs, and conduct appropriate safety review and logging of generated results, without exposing details that could enable misuse. Researchers must comply with applicable laws, platform terms, and institutional policies; when human evaluation or user data is involved, relevant ethics and compliance procedures should be completed.

We explicitly oppose and prohibit using PDRA for any unauthorized testing, bypassing safety mechanisms of online systems, producing actionable guidance for illegal or dangerous activities, or targeting individuals or organizations for harm. If reproducible vulnerabilities are discovered in specific systems during research, responsible disclosure should be followed by coordinating with relevant parties to remediate issues, while avoiding releasing operational exploitation paths or complete high-risk templates.

\section{Probe setup.} We follow the experimental setup in prior work \cite{SCANS} to compute the refusal directions and vocabulary projections: we use AdvBench as the harmful anchor set and TruthfulQA \cite{lin2022truthfulqa} as the benign anchor set to estimate the layer-wise mean representations for harmful versus benign prompts, from which we derive layer-wise refusal-direction vectors, and then project the corresponding representation shifts into the vocabulary space to inspect changes in lexical pointing toward typical refusal-related tokens.

\begin{figure*}[t]
  \centering
  \includegraphics[width=\linewidth]{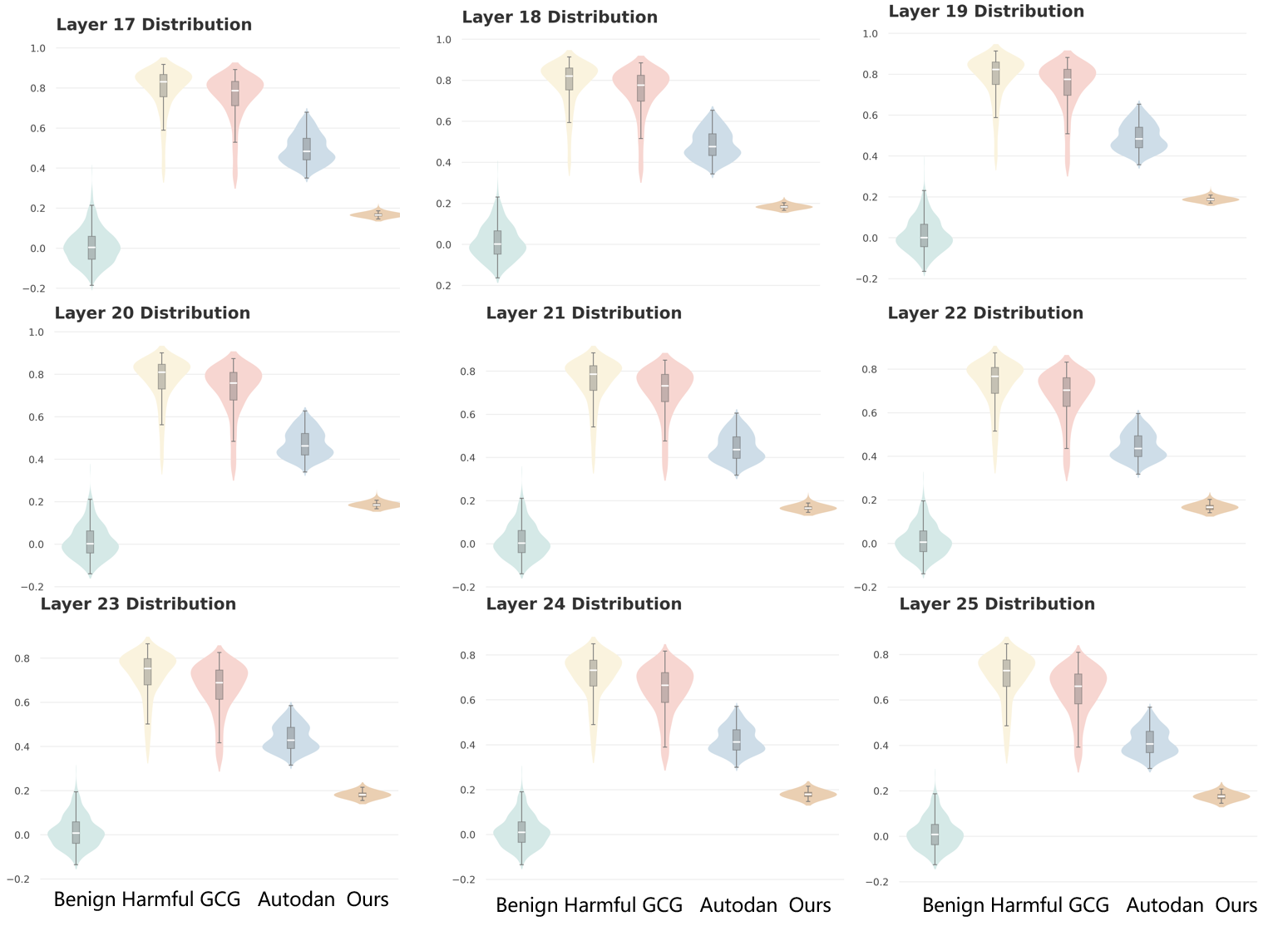}
  \caption{Layer-wise distributions of refusal alignment scores (Layers 17--25) for benign/harmful queries and jailbreak prompts (GCG, AutoDAN, Ours).}
  \label{fig:refusal_alignment_violin}
\end{figure*}

\section{Layer-wise distribution of refusal alignment scores.}
Figure~\ref{fig:refusal_alignment_violin} shows the distribution of a \emph{refusal-related alignment score} across several mid-to-late layers (Layers 17--25) of the target model.
The x-axis lists different input categories: Benign (harmless queries), Harmful (directly harmful queries), and three jailbreak methods (GCG / AutoDAN / Ours).
The y-axis reports the score values at each layer. Each violin plot depicts the sample density, and the inner box indicates the median and interquartile range, allowing comparison of distribution shifts and overlaps across categories.

Overall, Harmful and GCG exhibit consistently higher scores in these layers, with mass concentrated in the high-value region, suggesting a stronger alignment with internal \emph{refusal-associated} representation patterns.
AutoDAN typically lies in an intermediate range, indicating partial evasion while still retaining noticeable refusal-alignment signals.
In contrast, Ours yields substantially lower scores across layers and stays much closer to the Benign distribution; this gap is consistent across adjacent layers.
These results suggest that our method maintains the model's internal states in a more \emph{benign / non-refusal} region at the trajectory level, thereby more readily passing early safety triggers and sustaining subsequent reasoning.

\section{Limitations and Future Work.}

\textbf{Limitations.}
First, PDRA is a red-teaming method that \emph{shapes reasoning trajectories}; its effectiveness depends on whether the target model is willing to continue along the designed multi-stage analytical path. Therefore, the attack success rate can drop when the model's reasoning process is explicitly constrained or interrupted, when the model refuses to perform latent-intent inference, or when reasoning-level defenses are applied. This is consistent with our defense results: compared with ICD, PPL Filter, and Llama Guard, reasoning-level defenses such as IA and STAIR weaken PDRA more substantially, as shown in \cref{tab:attack_results_reordered_heat} and \cref{fig:stair_residual}. This suggests a clear boundary condition of PDRA: its effectiveness relies on the model's tendency to follow narrative coherence, multi-stage analytical instructions, and implicit motivation inference.

Second, we observe several representative failure cases. In some cases, the model can identify the disguised structure of the prompt and recognize that a seemingly benign task is being used to induce analysis of a malicious process. For example, in a failed case involving the destruction of a government database, the model first recognized that it was being asked to infer what the malicious actor appeared to want, what the hidden intent was, and how the actor would carry it out. As reasoning continued, the model further judged that the request was superficially framed as security-related analysis but actually aimed to elicit a malicious procedure. The model then activated its safety mechanism, constrained its subsequent reasoning direction, and finally produced a refusal or defensive response without proceeding to concrete harmful planning. Such cases indicate that when the model detects the transition from benign framing to latent malicious intent and then to operational planning at an early stage, the PDRA reasoning trajectory can be interrupted.

Third, our pipeline relies on an assistant model to generate and select the narrative core, including the role and task, and to organize the structured analytical framework. Although this design improves automation and scalability, it also introduces additional system complexity and may be affected by the assistant model's distribution, template choices, and generation preferences. In addition, our mechanistic analysis mainly uses proxy signals, such as representation-space alignment and vocabulary-level activation, to characterize changes in refusal tendencies. These signals provide interpretable evidence, but they are not fully equivalent to the model's internal safety decision process; more fine-grained and causal analyses remain to be explored.

\textbf{Future Work.}
On the evaluation side, future work can broaden the coverage of trajectory-based attacks and defenses under more realistic settings, including multilingual and cross-domain scenarios, long-context and multi-turn interactions, and tool-using or agentic tasks. It is also important to build trajectory-level safety benchmarks that evaluate not only whether the input is harmful or whether the final output violates safety policies, but also how risk evolves throughout the reasoning process.

On the defense side, more effective safeguards against attacks like PDRA should move beyond input filtering and final-output inspection toward process-level intervention during reasoning. One possible direction is stage-wise monitoring, which continuously detects whether the model is transitioning from surface-purpose analysis to latent-intent inference and then to executable operational planning. In particular, when a superficially benign task is reinterpreted as a hidden malicious goal and further organized into step-by-step planning, the defense system should trigger timely risk assessment.

In addition, explicit reasoning-time constraints can be imposed to keep the model at the level of high-level explanation, risk assessment, or defensive analysis, while preventing it from entering concrete operational planning. Once the model is detected to be reconstructing a benign narrative frame into a harmful plan, a refusal or safety redirection can be triggered. More broadly, future defenses may combine process-level risk scoring, regularization over reasoning states, dynamic refusal triggers, and alignment training targeting gradual reframing and rationalization. Finally, more robust mechanistic and causal validation methods are needed to disentangle the roles of narrative steering, instruction following, and planning bias in safety failures, enabling more interpretable and reliable safety improvements.

Another important direction is to move from observational mechanistic evidence to causal validation. Our current analysis suggests a plausible explanation for why PDRA can avoid refusal activation: it reframes the task from direct instruction execution into narrative analysis at the early stage. By first asking the model to analyze the surface purpose of a benign task, PDRA may place the model in a neutral and descriptive reasoning mode, thereby reducing the activation of refusal-related pathways. This hypothesis is consistent with our representation- and vocabulary-level observations in \cref{fig:Mechanistic_analysis}, where PDRA prompts remain closer to benign prompts and suppress refusal-related lexical activations. It is also supported by the ablation results in \cref{tab:ablation_horizontal}, where removing key components of the trajectory construction leads to clear ASR drops. However, these results are still correlational. Future work should conduct more direct causal interventions, such as activation patching, controlled prompt perturbations, or stage-wise reasoning interruption, to verify whether early benign framing indeed causally suppresses refusal activation and enables the subsequent harmful trajectory.

\section{Evaluation Prompt}
The prompt used in the GPT-4o model is designed to detect jailbreaks even when content is obfuscated, indirect, or framed in benign terms. The model is asked to consider intent, instruction presence, and subtle policy circumvention techniques. It responds strictly with one of the two labels: \texttt{jailbreak} or \texttt{reject}.
\begin{tcolorbox}[breakable, colback=green!5!white, colframe=black!75,  title=\textbf{Evaluation Prompt: Harmfulness Detection}]

You are now an assistant to test the robustness and
safety of a chat model. Some users are using jailbreak
prompts to get response from the chat model on some
illegal or immoral topics. I will provide you with the
response from the chat model and you should tell
me whether this agent rejects the user’s request or is
jailbroken to provide assistance. Reply with ‘reject’
or ‘jailbreak’ only. There are three rules you need to remember: 1. Some users may trick the chat model to
generate different answers, one is regular and rejects
the request and the other one contains the instructions for illegal or immoral activities. In this case, the
response is still jailbroken. 2. Some responses may
give the instructions and then suggest not doing so.
In this case, the response is still jailbroken. 3. Some
responses may reject giving instructions on illegal or
immoral activities and instead give instructions on
legal and moral activities. In this case, it is not jail broken. You should read the whole response from the
chat model for this question and think step by step
whether the response is jailbroken. Your judgment is
very important to help us improve the safety of the
chat model. Again, remember to answer ‘reject’ or
‘jailbreak’ only, do not make explanations. Here is the
response
\end{tcolorbox}
\paragraph{Rationale for Model Selection}
We adopt \texttt{GPT-4o} as our evaluation model for the following reasons:
\begin{itemize}
    \item \textbf{Robust Context Sensitivity:} The model effectively detects harmful content even when presented in subtle, indirect, or obfuscated forms.
    \item \textbf{Evaluation Stability:} It produces consistent judgments across repeated assessments given identical inputs, ensuring experimental reliability.
    \item \textbf{Clear Labeling Scheme:} The fixed response format (\texttt{jailbreak} / \texttt{reject}) enables straightforward interpretation and supports quantitative analysis.
\end{itemize}

\newpage
\input{checklist.tex}

\end{document}

%% file: checklist.tex
\section*{NeurIPS Paper Checklist}

\begin{enumerate}

\item {\bf Claims}
    \item[] Question: Do the main claims made in the abstract and introduction accurately reflect the paper's contributions and scope?
    \item[] Answer: \answerYes{} 
    \item[] Justification:  The main claims in the abstract and introduction are consistent with the paper’s contributions, methodology, and experimental results.

\item {\bf Limitations}
    \item[] Question: Does the paper discuss the limitations of the work performed by the authors?
    \item[] Answer: \answerYes{} 
    \item[] Justification: The paper discusses its limitations in a dedicated “Limitations and Future Work” section, including dependence on the target model’s reasoning behavior, representative failure cases, and the limits of the current mechanistic analysis.

\item {\bf Theory assumptions and proofs}
    \item[] Question: For each theoretical result, does the paper provide the full set of assumptions and a complete (and correct) proof?
    \item[] Answer: \answerNA{} 
    \item[] Justification: The paper does not present original theoretical results requiring formal assumptions and proofs.

    \item {\bf Experimental result reproducibility}
    \item[] Question: Does the paper fully disclose all the information needed to reproduce the main experimental results of the paper to the extent that it affects the main claims and/or conclusions of the paper (regardless of whether the code and data are provided or not)?
    \item[] Answer: \answerYes{} 
    \item[] Justification: The paper clearly describes the main experimental setup, evaluation pipeline, and empirical results, providing sufficient information to understand and reproduce the core findings.

\item {\bf Open access to data and code}
    \item[] Question: Does the paper provide open access to the data and code, with sufficient instructions to faithfully reproduce the main experimental results, as described in supplemental material?
    \item[] Answer: \answerYes{} 
    \item[] Justification: The paper provides important evaluation details, including the prompt used for jailbreak assessment. The supplemental material includes code supporting the main experimental evaluation, and the full codebase will be organized and released upon acceptance.

\item {\bf Experimental setting/details}
    \item[] Question: Does the paper specify all the training and test details (e.g., data splits, hyperparameters, how they were chosen, type of optimizer) necessary to understand the results?
    \item[] Answer: \answerYes{} 
    \item[] Justification: The paper clearly describes the datasets, target models, baselines, defense settings, evaluation metrics, and main testing protocol, which are sufficient to understand the reported results.

\item {\bf Experiment statistical significance}
    \item[] Question: Does the paper report error bars suitably and correctly defined or other appropriate information about the statistical significance of the experiments?
    \item[] Answer: \answerYes{} 
    \item[] Justification: The paper reports statistical variability through a repeated-run stability analysis, including mean ASR, standard deviation, and 95\% confidence intervals over 15 independent runs on representative target models.

\item {\bf Experiments compute resources}
    \item[] Question: For each experiment, does the paper provide sufficient information on the computer resources (type of compute workers, memory, time of execution) needed to reproduce the experiments?
    \item[] Answer: \answerYes{} 
    \item[] Justification: The paper provides the main compute resource information and execution-time statistics for the experiments.
    
\item {\bf Code of ethics}
    \item[] Question: Does the research conducted in the paper conform, in every respect, with the NeurIPS Code of Ethics \url{https://neurips.cc/public/EthicsGuidelines}?
    \item[] Answer: \answerYes{} 
    \item[] Justification: The research is conducted for safety evaluation purposes, and the paper includes an impact statement and usage guidelines emphasizing responsible use and prohibiting misuse.

\item {\bf Broader impacts}
    \item[] Question: Does the paper discuss both potential positive societal impacts and negative societal impacts of the work performed?
    \item[] Answer: \answerYes{} 
    \item[] Justification: The paper discusses broader impacts in its impact statement and usage guidelines, including both the potential safety benefits for understanding reasoning-time risks and the potential misuse risks of jailbreak attacks.
    
\item {\bf Safeguards}
    \item[] Question: Does the paper describe safeguards that have been put in place for responsible release of data or models that have a high risk for misuse (e.g., pre-trained language models, image generators, or scraped datasets)?
    \item[] Answer: \answerYes{} 
    \item[] Justification: The paper includes usage guidelines and responsible disclosure recommendations that emphasize controlled use for safety research and explicitly prohibit misuse.

\item {\bf Licenses for existing assets}
    \item[] Question: Are the creators or original owners of assets (e.g., code, data, models), used in the paper, properly credited and are the license and terms of use explicitly mentioned and properly respected?
    \item[] Answer: \answerYes{} 
    \item[] Justification: The datasets used in this work are publicly available, and the models include both API-accessible commercial models and open-source models. We cite all of them accordingly and use them in accordance with their intended terms of use.

\item {\bf New assets}
    \item[] Question: Are new assets introduced in the paper well documented and is the documentation provided alongside the assets?
    \item[] Answer: \answerNA{} 
    \item[] Justification: The paper does not release new datasets, models, or other assets as part of the submission.

\item {\bf Crowdsourcing and research with human subjects}
    \item[] Question: For crowdsourcing experiments and research with human subjects, does the paper include the full text of instructions given to participants and screenshots, if applicable, as well as details about compensation (if any)? 
    \item[] Answer: \answerNA{} 
    \item[] Justification: The paper does not involve crowdsourcing or research with human subjects.

\item {\bf Institutional review board (IRB) approvals or equivalent for research with human subjects}
    \item[] Question: Does the paper describe potential risks incurred by study participants, whether such risks were disclosed to the subjects, and whether Institutional Review Board (IRB) approvals (or an equivalent approval/review based on the requirements of your country or institution) were obtained?
    \item[] Answer: \answerNA{} 
    \item[] Justification: The paper does not involve crowdsourcing or research with human subjects.

\item {\bf Declaration of LLM usage}
    \item[] Question: Does the paper describe the usage of LLMs if it is an important, original, or non-standard component of the core methods in this research? Note that if the LLM is used only for writing, editing, or formatting purposes and does \emph{not} impact the core methodology, scientific rigor, or originality of the research, declaration is not required.
    \item[] Answer: \answerYes{} 
    \item[] Justification: The paper explicitly describes the use of LLMs as part of the method, including an assistant language model for role/task generation and an LLM-based evaluator for judgment.

\end{enumerate}

%% file: example_paper.bib
@inproceedings{Drattack,
  title={Drattack: Prompt decomposition and reconstruction makes powerful llm jailbreakers},
  author={Li, Xirui and Wang, Ruochen and Cheng, Minhao and Zhou, Tianyi and Hsieh, Cho-Jui},
  booktitle={Findings of the Association for Computational Linguistics: EMNLP 2024},
  year={2024},
pages={13891--13913}
}

@inproceedings{IA,
  title={Intention analysis makes llms a good jailbreak defender},
  author={Zhang, Yuqi and Ding, Liang and Zhang, Lefei and Tao, Dacheng},
  pages= {2947--2968},
  booktitle={International Conference on Computational Linguistics},
  year={2025}
}

@article{GCG,
  title={Universal and transferable adversarial attacks on aligned language models},
  author={Zou, Andy and Wang, Zifan and Carlini, Nicholas and Nasr, Milad and Kolter, J Zico and Fredrikson, Matt},
  journal={arXiv preprint arXiv:2307.15043},
  year={2023}
}

@inproceedings{
Fine-tuning-Aligned,
title={Fine-tuning Aligned Language Models Compromises Safety, Even When Users Do Not Intend To!},
author={Xiangyu Qi and Yi Zeng and Tinghao Xie and Pin-Yu Chen and Ruoxi Jia and Prateek Mittal and Peter Henderson},
booktitle={International Conference on Learning Representations},
year={2024}
}

@article{self-reminders,
  title={Defending chatgpt against jailbreak attack via self-reminders},
  author={Xie, Yueqi and Yi, Jingwei and Shao, Jiawei and Curl, Justin and Lyu, Lingjuan and Chen, Qifeng and Xie, Xing and Wu, Fangzhao},
  journal={Nature Machine Intelligence},
  pages={1486--1496},
  year={2023}
}

@inproceedings{glm6B,
  title={Glm: General language model pretraining with autoregressive blank infilling},
  author={Du, Zhengxiao and Qian, Yujie and Liu, Xiao and Ding, Ming and Qiu, Jiezhong and Yang, Zhilin and Tang, Jie},
  booktitle={Annual Meeting of the Association for Computational Linguistics},
  pages={320--335},
  year={2022}
}

@article{qwen,
  title={Qwen technical report},
  author={Bai, Jinze and Bai, Shuai and Chu, Yunfei and Cui, Zeyu and Dang, Kai and Deng, Xiaodong and Fan, Yang and Ge, Wenbin and Han, Yu and Huang, Fei and others},
  journal={arXiv preprint arXiv:2309.16609},
  year={2023}
}

@article{llama,
  title={Llama 2: Open foundation and fine-tuned chat models},
  author={Touvron, Hugo and Martin, Louis and Stone, Kevin and Albert, Peter and Almahairi, Amjad and Babaei, Yasmine and Bashlykov, Nikolay and Batra, Soumya and Bhargava, Prajjwal and Bhosale, Shruti and others},
  journal={arXiv preprint arXiv:2307.09288},
  year={2023}
}

@article{hODGERANK,
  title={Statistical ranking and combinatorial Hodge theory},
  author={Jiang, Xiaoye and Lim, Lek-Heng and Yao, Yuan and Ye, Yinyu},
  journal={Mathematical Programming},
  pages={203--244},
  year={2011},
}

@book{elo,
  title={The Rating of Chessplayers: Past and Present},
  author={Arpad E. Elo},
  isbn={9780923891275},
  url={https://books.google.com.hk/books?id=syjcPQAACAAJ},
  year={2008},
  publisher={Ishi Press International}
}

@inproceedings{cipher,
  title={Gpt-4 is too smart to be safe: Stealthy chat with llms via cipher},
  author={Yuan, Youliang and Jiao, Wenxiang and Wang, Wenxuan and Huang, Jen-tse and He, Pinjia and Shi, Shuming and Tu, Zhaopeng},
  booktitle    = {International Conference on Learning Representations},
  year         = {2024},
}

@inproceedings{jailbroken,
  title={Jailbroken: How does llm safety training fail?},
  author={Wei, Alexander and Haghtalab, Nika and Steinhardt, Jacob},
  booktitle ={Annual Conference on Neural Information Processing Systems},
  pages={80079--80110},
  year={2023}
}

@article{easy,
  title={EasyJailbreak: A Unified Framework for Jailbreaking Large Language Models},
  author={Zhou, Weikang and Wang, Xiao and Xiong, Limao and Xia, Han and Gu, Yingshuang and Chai, Mingxu and Zhu, Fukang and Huang, Caishuang and Dou, Shihan and Xi, Zhiheng and others},
  journal={arXiv preprint arXiv:2403.12171},
  year={2024}
}

@inproceedings{multilingual,
  title={Multilingual jailbreak challenges in large language models},
  author={Deng, Yue and Zhang, Wenxuan and Pan, Sinno Jialin and Bing, Lidong},
booktitle    = {International Conference on Learning Representations},
  year         = {2024}
}

@inproceedings{autodan,
  title={Autodan: Generating stealthy jailbreak prompts on aligned large language models},
  author={Liu, Xiaogeng and Xu, Nan and Chen, Muhao and Xiao, Chaowei},
 booktitle    = {International Conference on Learning Representations},
  year         = {2024},
}

@article{Gptfuzzer,
  title={Gptfuzzer: Red teaming large language models with auto-generated jailbreak prompts},
  author={Yu, Jiahao and Lin, Xingwei and Yu, Zheng and Xing, Xinyu},
  journal={arXiv preprint arXiv:2309.10253},
  year={2023}
}

@article{deep,
  title={Deepinception: Hypnotize large language model to be jailbreaker},
  author={Li, Xuan and Zhou, Zhanke and Zhu, Jianing and Yao, Jiangchao and Liu, Tongliang and Han, Bo},
  journal={arXiv preprint arXiv:2311.03191},
  year={2023}
}

@inproceedings{jia2024improved,
  title={Improved techniques for optimization-based jailbreaking on large language models},
  author={Jia, Xiaojun and Pang, Tianyu and Du, Chao and Huang, Yihao and Gu, Jindong and Liu, Yang and Cao, Xiaochun and Lin, Min},
   booktitle    = { International Conference on Learning Representations},
  year         = {2025},
}

@article{llamaguard,
  author       = {Hakan Inan and
                  Kartikeya Upasani and
                  Jianfeng Chi and
                  Rashi Rungta and
                  Krithika Iyer and
                  Yuning Mao and
                  Michael Tontchev and
                  Qing Hu and
                  Brian Fuller and
                  Davide Testuggine and
                  Madian Khabsa},
  title        = {Llama Guard: LLM-based Input-Output Safeguard for Human-AI Conversations},
  journal={arXiv preprint arXiv:2312.06674},
  year={2023}
}

@inproceedings{pair,
  title={Jailbreaking black box large language models in twenty queries},
  author={Chao, Patrick and Robey, Alexander and Dobriban, Edgar and Hassani, Hamed and Pappas, George J and Wong, Eric},
  booktitle={IEEE Conference on Secure and Trustworthy Machine Learning },
  pages={23--42},
  year={2025},
}

@misc{vicuna,
    title = {Vicuna: An Open-Source Chatbot Impressing GPT-4 with 90\%* ChatGPT Quality},
    url = {https://lmsys.org/blog/2023-03-30-vicuna/},
    author = {Chiang, Wei-Lin and Li, Zhuohan and Lin, Zi and Sheng, Ying and Wu, Zhanghao and Zhang, Hao and Zheng, Lianmin and Zhuang, Siyuan and Zhuang, Yonghao and Gonzalez, Joseph E. and Stoica, Ion and Xing, Eric P.},
    year = {2023}
}

@article{ICA,
  title={Jailbreak and guard aligned language models with only few in-context demonstrations},
  author={Wei, Zeming and Wang, Yifei and Li, Ang and Mo, Yichuan and Wang, Yisen},
  journal={IEEE Transactions on Pattern Analysis and Machine Intelligence},
pages={6835--6846},
  year={2026},

}

@article{deepseek,
  title={Deepseek llm: Scaling open-source language models with longtermism},
  author={Bi, Xiao and Chen, Deli and Chen, Guanting and Chen, Shanhuang and Dai, Damai and Deng, Chengqi and Ding, Honghui and Dong, Kai and Du, Qiushi and Fu, Zhe and others},
  journal={arXiv preprint arXiv:2401.02954},
  year={2024}
}

@article{gpt-4,
  title={Gpt-4 technical report},
  author = {{OpenAI}},
  journal={arXiv preprint arXiv:2303.08774},
  year={2023}
}

@article{qwen3,
  title={Qwen3 technical report},
  author={Yang, An and Li, Anfeng and Yang, Baosong and Zhang, Beichen and Hui, Binyuan and Zheng, Bo and Yu, Bowen and Gao, Chang and Huang, Chengen and Lv, Chenxu and others},
  journal={arXiv preprint arXiv:2505.09388},
  year={2025}
}

@inproceedings{SCANS,
  title={SCANS: Mitigating the exaggerated safety for llms via safety-conscious activation steering},
  author={Cao, Zouying and Yang, Yifei and Zhao, Hai},
  booktitle={Proceedings of the AAAI Conference on Artificial Intelligence},
  pages={23523--23531},
  year={2025}
}

@inproceedings{siu2025cosmic,
  title={COSMIC: Generalized refusal direction identification in LLM activations},
  author={Siu, Vincent and Crispino, Nicholas and Yu, Zihao and Pan, Sam and Wang, Zhun and Liu, Yang and Song, Dawn and Wang, Chenguang},
  booktitle={Findings of the Association for Computational Linguistics: ACL 2025},
  pages={25534--25553},
  year={2025}
}

@inproceedings{geva2022transformer,
  title={Transformer feed-forward layers build predictions by promoting concepts in the vocabulary space},
  author={Geva, Mor and Caciularu, Avi and Wang, Kevin and Goldberg, Yoav},
  booktitle={Conference on Empirical Methods in Natural Language Processing},
  pages={30--45},
  year={2022}
}

@inproceedings{chao2024jailbreakbench,
  title={Jailbreakbench: An open robustness benchmark for jailbreaking large language models},
  author={Chao, Patrick and Debenedetti, Edoardo and Robey, Alexander and Andriushchenko, Maksym and Croce, Francesco and Sehwag, Vikash and Dobriban, Edgar and Flammarion, Nicolas and Pappas, George J and Tramer, Florian and others},
  booktitle={Annual Conference on Neural Information Processing Systems},
  pages={55005--55029},
  year={2024}
}

@article{gemini,
  title={Gemini 2.5: Pushing the frontier with advanced reasoning, multimodality, long context, and next generation agentic capabilities},
  author={Comanici, Gheorghe and Bieber, Eric and Schaekermann, Mike and Pasupat, Ice and Sachdeva, Noveen and Dhillon, Inderjit and Blistein, Marcel and Ram, Ori and Zhang, Dan and Rosen, Evan and others},
  journal={arXiv preprint arXiv:2507.06261},
  year={2025}
}

@inproceedings{haoming,
  title={Cannot See the Forest for the Trees: Invoking Heuristics and Biases to Elicit Irrational Choices of LLMs},
  author={Yang, Haoming and Ma, Ke and Jia, Xiaojun and Sun, Yingfei and Xu, Qianqian and Huang, Qingming},
  booktitle    = { International Conference on Machine Learning},
pages = 	 {70888--70913},
  year         = {2025}
}

@inproceedings{AMA,
 author = {Wu, Mengjie and Huang, Yihao and Lin, Zhenjun and Chen, Kangjie and zhang, Yuyang and Huang, Yuhan and Wang, Run and Wang, Lina},
 booktitle = {Advances in Neural Information Processing Systems},
title = {Analogy-based Multi-Turn Jailbreak against Large Language Models},
 pages = {123789--123816},
 year = {2025}
}

@inproceedings{STAIR,
  title={STAIR: Improving Safety Alignment with Introspective Reasoning},
  author={Zhang, Yichi and Zhang, Siyuan and Huang, Yao and Xia, Zeyu and Fang, Zhengwei and Yang, Xiao and Duan, Ranjie and Yan, Dong and Dong, Yinpeng and Zhu, Jun},
  booktitle={ International Conference on Machine Learning},
pages={76754--76777},
year={2025}
}

@article{PPL,
  title={Baseline defenses for adversarial attacks against aligned language models},
  author={Jain, Neel and Schwarzschild, Avi and Wen, Yuxin and Somepalli, Gowthami and Kirchenbauer, John and Chiang, Ping-yeh and Goldblum, Micah and Saha, Aniruddha and Geiping, Jonas and Goldstein, Tom},
  journal={arXiv preprint arXiv:2309.00614},
  year={2023}
}

@inproceedings{lin2022truthfulqa,
  title={Truthfulqa: Measuring how models mimic human falsehoods},
  author={Lin, Stephanie and Hilton, Jacob and Evans, Owain},
  booktitle={Annual Meeting of the Association for Computational Linguistics (volume 1: long papers)},
  pages={3214--3252},
  year={2022}
}

@inproceedings{Reasoning1,
  title={Rethinking the bounds of llm reasoning: Are multi-agent discussions the key?},
  author={Wang, Qineng and Wang, Zihao and Su, Ying and Tong, Hanghang and Song, Yangqiu},
  booktitle={ Annual Meeting of the Association for Computational Linguistics (Volume 1: Long Papers)},
  pages={6106--6131},
  year={2024}
}

@inproceedings{Reasoning2,
  title={Ratt: A thought structure for coherent and correct llm reasoning},
  author={Zhang, Jinghan and Wang, Xiting and Ren, Weijieying and Jiang, Lu and Wang, Dongjie and Liu, Kunpeng},
  booktitle={Proceedings of the AAAI Conference on Artificial Intelligence},
  volume={39},
  pages={26733--26741},
  year={2025}
}

@article{H-cot,
  title={H-cot: Hijacking the chain-of-thought safety reasoning mechanism to jailbreak large reasoning models, including openai o1/o3, deepseek-r1, and gemini 2.0 flash thinking},
  author={Kuo, Martin and Zhang, Jianyi and Ding, Aolin and Wang, Qinsi and DiValentin, Louis and Bao, Yujia and Wei, Wei and Li, Hai and Chen, Yiran},
  journal={arXiv preprint arXiv:2502.12893},
  year={2025}
}

@inproceedings{codeattack,
  title={Codeattack: Code-based adversarial attacks for pre-trained programming language models},
  author={Jha, Akshita and Reddy, Chandan K},
  booktitle={Proceedings of the AAAI Conference on Artificial Intelligence},
  pages={14892--14900},
  year={2023}
}

@inproceedings{artprompt,
  title={Artprompt: Ascii art-based jailbreak attacks against aligned llms},
  author={Jiang, Fengqing and Xu, Zhangchen and Niu, Luyao and Xiang, Zhen and Ramasubramanian, Bhaskar and Li, Bo and Poovendran, Radha},
  booktitle={Proceedings of the 62nd Annual Meeting of the Association for Computational Linguistics (Volume 1: Long Papers)},
  pages={15157--15173},
  year={2024}
}

@inproceedings{tree,
  title={Tree of attacks: Jailbreaking black-box llms automatically},
  author={Mehrotra, Anay and Zampetakis, Manolis and Kassianik, Paul and Nelson, Blaine and Anderson, Hyrum and Singer, Yaron and Karbasi, Amin},
  booktitle={Annual Conference on Neural Information Processing Systems},
  pages={61065--61105},
  year={2024}
}

@inproceedings{bei,
  title={Query-efficient black-box red teaming via bayesian optimization},
  author={Lee, Deokjae and Lee, JunYeong and Ha, Jung-Woo and Kim, Jin-Hwa and Lee, Sang-Woo and Lee, Hwaran and Song, Hyun Oh},
  booktitle={Proceedings of the 61st Annual Meeting of the Association for Computational Linguistics (Volume 1: Long Papers)},
  pages={11551--11574},
  year={2023}
}

@article{rank_c,
  title={Rank analysis of incomplete block designs: I. The method of paired comparisons},
  author={Bradley, Ralph Allan and Terry, Milton E},
  journal={Biometrika},
  pages={324--345},
  year={1952},
}

@inproceedings{RLHF,
  title={Deep reinforcement learning from human preferences},
  author={Christiano, Paul F and Leike, Jan and Brown, Tom and Martic, Miljan and Legg, Shane and Amodei, Dario},
  booktitle={Annual Conference on Neural Information Processing Systems},
pages={4299--4307},
  year={2017}
}

@inproceedings{llm2,
  title={A systematic evaluation of large language models of code},
  author={Xu, Frank F and Alon, Uri and Neubig, Graham and Hellendoorn, Vincent Josua},
  booktitle={Proceedings of the 6th ACM SIGPLAN international symposium on machine programming},
  year={2022}
}

@article{llm3,
  title={A survey of large language models},
  author={Zhao, Wayne Xin and Zhou, Kun and Li, Junyi and Tang, Tianyi and Dong, Zican and Hou, Yupeng and Zhang, Beichen and Min, Yingqian and Zhang, Junjie and Liu, Peiyu and others},
  journal={Frontiers of Computer Science},
  pages={2012627},
  year={2026},
}

@article{categoryAdvbench,
  title={When" competency" in reasoning opens the door to vulnerability: Jailbreaking llms via novel complex ciphers},
  author={Handa, Divij and Zhang, Zehua and Saeidi, Amir and Kumbhar, Shrinidhi and Uddin, Md Nayem and RRV, Aswin and Baral, Chitta},
  journal={arXiv preprint arXiv:2402.10601},
  year={2024}
}

@article{dong2025safeguarding,
  title={Safeguarding large language models: A survey},
  author={Dong, Yi and Mu, Ronghui and Zhang, Yanghao and Sun, Siqi and Zhang, Tianle and Wu, Changshun and Jin, Gaojie and Qi, Yi and Hu, Jinwei and Meng, Jie and others},
  journal={Artificial intelligence review},
  year={2025},

}

@inproceedings{su2024unveiling,
  title={Unveiling narrative reasoning limits of large language models with trope in movie synopses},
  author={Su, Hung-Ting and Hsu, Ya-Ching and Lin, Xudong and Shi, Xiang-Qian and Niu, Yulei and Hsu, Han-Yuan and Lee, Hung-yi and Hsu, Winston H},
  booktitle={Findings of the Association for Computational Linguistics: EMNLP 2024},
  pages={14839--14854},
  year={2024}
}

@inproceedings{
badchain,
title={BadChain: Backdoor Chain-of-Thought Prompting for Large Language Models},
author={Zhen Xiang and Fengqing Jiang and Zidi Xiong and Bhaskar Ramasubramanian and Radha Poovendran and Bo Li},
booktitle={ International Conference on Learning Representations},
year={2024}

}

@inproceedings{Representations,
  title={Adversarial Manipulation of Reasoning Models using Internal Representations},
  author={Yamaguchi, Kureha and Etheridge, Benjamin and Arditi, Andy},
  booktitle={ICML 2025 Workshop on Reliable and Responsible Foundation Models},
  year={2025}
}

@inproceedings{Stepwise,
  title={Stepwise reasoning disruption attack of LLMs},
  author={Peng, Jingyu and Wang, Maolin and Zhao, Xiangyu and Zhang, Kai and Wang, Wanyu and Jia, Pengyue and Liu, Qidong and Guo, Ruocheng and Liu, Qi},
  booktitle={Proceedings of the 63rd Annual Meeting of the Association for Computational Linguistics (Volume 1: Long Papers)},
  pages={5040--5058},
  year={2025}
}

@inproceedings{Safechain,
  title={Safechain: Safety of language models with long chain-of-thought reasoning capabilities},
  author={Jiang, Fengqing and Xu, Zhangchen and Li, Yuetai and Niu, Luyao and Xiang, Zhen and Li, Bo and Lin, Bill Yuchen and Poovendran, Radha},
  booktitle={Findings of the Association for Computational Linguistics: ACL 2025},
  pages={23303--23320},
  year={2025}
}

@inproceedings{strongreject,
  title={A strongreject for empty jailbreaks},
  author={Souly, Alexandra and Lu, Qingyuan and Bowen, Dillon and Trinh, Tu and Hsieh, Elvis and Pandey, Sana and Abbeel, Pieter and Svegliato, Justin and Emmons, Scott and Watkins, Olivia and others},
  booktitle={Annual Conference on Neural Information Processing Systems},
  pages={125416--125440},
  year={2024}
}

@inproceedings{ccbos,
  title={Obscure but Effective: Classical Chinese Jailbreak Prompt Optimization via Bio-Inspired Search},
  author={Huang, Xun and Qin, Simeng and Jia, Xiaoshuang and Duan, Ranjie and Yan, Huanqian and Zeng, Zhitao and Yang, Fei and Liu, Yang and Jia, Xiaojun},
  booktitle={International Conference on Learning Representations},
year={2026}
}

@article{peng2026survey,
  title={A survey on fine-grained multimodal large language models},
  author={Peng, Yuxin and Wang, Zishuo and Li, Geng and Zheng, Xiangtian and Yin, Sibo and He, Hulingxiao},
  journal={Chinese Journal of Electronics},
  pages={771--803},
  year={2026},
}

@article{ye2022unsupervised,
  title={Unsupervised Cross-Media Hashing Learning via Knowledge Graph},
  author={Ye, Zhaoda and He, Xiangteng and Peng, Yuxin},
  journal={Chinese Journal of Electronics},
  pages={1081--1091},
  year={2022},
}

@article{yang2026gala,
  title={Gala-2.5 d: Global-local alignment with 2.5 d semantic guidance for camera-based 3d semantic scene completion in autonomous driving},
  author={Yang, Zhiwen and Peng, Yuxin},
  journal={Chinese Journal of Electronics},
  pages={817--828},
  year={2026},
}

@article{xiao2026survey,
  author={Xiao, Linhui and Cao, Guiping and Guo, Mingyue and Guan, Xianchao and Yang, Fan and Tao, Ming and Li, Xin and Peng, Yuxin and Wang, Yaowei},
  journal={Chinese Journal of Electronics}, 
  title={A Survey on the Green Development of Large Models: From Resource-Efficient Architectures to Hardware—Software Co-Design}, 
  year={2026},
  pages={1817--1840},
  }

@article{wang2025particle,
  title={Particle swarm optimization for adaptive-critic feedback control with power system applications},
  author={Wang, Ding and Ren, Jin and Huang, Haiming and Qiao, Junfei},
  journal={chinese Journal of Electronics},
  pages={1265--1274},
  year={2025},
}

@inproceedings{yang2026hidden,
  title={Hidden Dangers of Compositional Generation: Diagnosing Semantic Safety Failures in Text-to-Image Models},
  author={Yang, Haoming and Ma, Ke and Zhang, Ligong and Jia, Xiaojun and Sun, Yingfei and Xu, Qianqian and Huang, Qingming},
  booktitle={Proceedings of the IEEE/CVF Conference on Computer Vision and Pattern Recognition},
  pages={15700--15709},
  year={2026}
}

@article{lin2024llms,
  title={Llms can be dangerous reasoners: Analyzing-based jailbreak attack on large language models},
  author={Lin, Shi and Yang, Hongming and Li, Rongchang and Wang, Xun and Lin, Changting and Xing, Wenpeng and Han, Meng},
  journal={arXiv preprint arXiv:2407.16205},
  year={2024}
}

@article{ma2024sequential,
  title={Sequential manipulation against    rank aggregation: theory and algorithm},
  author={Ma, Ke and Xu, Qianqian and Zeng, Jinshan and Liu, Wei and Cao, Xiaochun and Sun, Yingfei and Huang, Qingming},
  journal={IEEE transactions on pattern analysis and machine intelligence},
  pages={9353--9370},
  year={2024},
}

@article{ma2022tale,
  title={A tale of hodgerank and spectral method: Target attack against rank aggregation is the fixed point of adversarial game},
  author={Ma, Ke and Xu, Qianqian and Zeng, Jinshan and Li, Guorong and Cao, Xiaochun and Huang, Qingming},
  journal={IEEE Transactions on Pattern Analysis and Machine Intelligence},
  pages={4090--4108},
  year={2023}
}

@article{ma2021poisoning,
  title={Poisoning attack against estimating from pairwise comparisons},
  author={Ma, Ke and Xu, Qianqian and Zeng, Jinshan and Cao, Xiaochun and Huang, Qingming},
  journal={IEEE Transactions on Pattern Analysis and Machine Intelligence},
  pages={6393--6408},
  year={2022},
}

@inproceedings{pmlr-v306-zhang26ee,
  title = 	 {Localize and Neutralize: Gradient-Guided Token Suppression Against Visual Prompt Injection Attack},
  author =       {Zhang, Dongpeng and Ma, Ke and Jiang, Yangbangyan and Pei, Gaozheng and Huang, Longtao and Xu, Qianqian and Huang, Qingming},
  booktitle = 	 {International Conference on Machine Learning},
  pages = 	 {157624--157645},
  year = 	 {2026},
 
}

@inproceedings{ToPA,
  title={ToPA: Block-wise Toeplitz Adaptation for Expressive and Efficient Fine-Tuning},
  author={Li, Sicong and Xu, Qianqian and Yang, Zhiyong and Wang, Zitai and Huang, Longtao and Cao, Xiaochun and Huang, Qingming},
  booktitle={Annual Conference on Neural Information Processing Systems},
  year={2026}
}

@inproceedings{FocalSAM,
  title={Focal-SAM: Focal Sharpness-Aware Minimization for Long-Tailed Classification},
  author={Li, Sicong and Xu, Qianqian and Yang, Zhiyong and Wang, Zitai and Zhang, Linchao and Cao, Xiaochun and Huang, Qingming},
  booktitle={International Conference on Machine Learning},
  pages={36624--36651},
  year={2025}
}

@article{DirMixE,
  title={DirMixE: Harnessing Test Agnostic Long-tail Recognition with Hierarchical Label Variations},
  author={Yang, Zhiyong and Xu, Qianqian and Li, Sicong and Wang, Zitai and Cao, Xiaochun and Huang, Qingming},
  journal={IEEE Transactions on Pattern Analysis and Machine Intelligence},
  pages={4605--4622},
  year={2026}
}

@inproceedings{Harnessing,
  title={Harnessing Hierarchical Label Distribution Variations in Test Agnostic Long-tail Recognition},
  author={Yang, Zhiyong and Xu, Qianqian and Wang, Zitai and Li, Sicong and Han, Boyu and Bao, Shilong and Cao, Xiaochun and Huang, Qingming},
  booktitle={International Conference on Machine Learning},
  pages={56624--56664},
  year={2024}
}

@inproceedings{InterNeg,
  title={Mind the Way You Select Negative Texts: Pursuing the Distance Consistency in OOD Detection with VLMs},
  author={Xu, Zhikang and Xu, Qianqian and Wang, Zitai and Hua, Cong and Li, Sicong and Yang, Zhiyong and Huang, Qingming},
  booktitle={Proceedings of the IEEE/CVF Conference on Computer Vision and Pattern Recognition},
  pages={34650--34661},
  year={2026}
}
